\documentclass[11pt,letterpaper]{article}
\usepackage[utf8]{inputenc}

\usepackage{amsmath}
\usepackage{graphicx}
\usepackage{subfig}
\usepackage{url}
\usepackage{hyperref}

\usepackage{amsmath,amssymb,amsfonts}
\usepackage{textcomp}
\usepackage{multirow}
\usepackage{multicol}
\usepackage{color}

\usepackage[numbers]{natbib}
\usepackage{caption}

\usepackage{booktabs}
\usepackage{tikz}
\usetikzlibrary{arrows,positioning}
\usepackage{lipsum, adjustbox}

\newcommand*\samethanks[1][\value{footnote}]{\footnotemark[#1]}

\begin{document}

\author{
\and Ozan Ozyegen
\thanks{Department of Mechanical and Industrial Engineering, Ryerson University, email: oozyegen@ryerson.ca}
\and Sanaz Mohammadjafari\samethanks[1]
\and Mucahit Cevik\samethanks[1]
\and Karim El mokhtari\samethanks[1]
\and Jonathan Ethier
\thanks{Communications Research Centre Canada}
\and Ayse Basar\samethanks[1]
}
\date{}

\title{An empirical study on using CNNs for fast radio signal prediction}
\maketitle


\begin{abstract}
Accurate radio frequency power prediction in a geographic region is a computationally expensive part of finding the optimal transmitter location using a ray tracing software. We empirically analyze the viability of deep learning models to speed up this process. Specifically, deep learning methods including CNNs and UNET are typically used for segmentation, and can also be employed in power prediction tasks. 
We consider a dataset that consists of radio frequency power values for five different regions with four different frame dimensions. 
We compare deep learning-based prediction models including RadioUNET and four different variations of the UNET model for the power prediction task.
More complex UNET variations improve the model on higher resolution frames such as 256$\times$256. 
However, using the same models on lower resolutions results in overfitting and simpler models perform better. 
Our detailed numerical analysis shows that the deep learning models are effective in power prediction and they are able to generalize well to the new regions.
\end{abstract}



\section{Introduction}
Predicting power coverage in urban areas typically requires a ray-tracing simulations to determine how radio signals propagate and are distributed over an area~\citep{insite}. 
Obstacles in the line of sight between any point and the transmitter can attenuate the propagated energy. 
Buildings and other objects can also cause interference, reflection and refraction which lead to complex energy coverage patterns that are difficult to determine accurately without ray-tracing. 
However, the ray-tracing process requires significant computational resources and a long simulation time as it applies propagation equations at every pixel in a given region. 

During the design phase of radio transmitter placement, one objective of ray-tracing is to determine a location for the transmitter. 
When placed in an optimal location, the transmitter power can reach a maximum number of points in the area leading to an important cost reduction in terms of numbers of transmitters to provide maximum coverage.
Ray-tracing simulation time increases proportionally to the region size and may extend the design phase considerably. 
In complex urban environments, it is not intuitive to determine whether a candidate location is optimal even with prior expertise in the field. 

In this study, we consider deep learning models to learn from ray-tracing results, and predict the power received at every point in a given area with one transmitter. 
Once the model is trained, the duration of the design phase can be reduced drastically since predicting the power coverage given a transmitter location in a communications environment can be done instantaneously. Specifically, we can determine a set of candidate optimal transmitter locations as a coarse approximation by comparing power coverage map predictions. 
Designers can then apply ray-tracing as a final fine-tuning step in the design process to choose the best location from the candidate set.

A deep learning-based prediction model work in 2D, and is fed with the building layouts and the transmitter location as a two separate binary input layers. 
The model is trained to produce the power value at every point in the region. 
The underlying task considered in this paper is similar to the semantic segmentation with the difference that the predicted values in segmentation are discrete and specify the class of the object in every pixel, while in our problem, the values are real numbers representing the signal power. 
We investigate using Convolutional Neural Networks (CNN) that was shown to learn efficiently from images, and UNET \citep{ronn2015} which is an extension of CNNs and a popular model in image segmentation. 
The encoder module of the UNET is used to capture the relation between buildings/transmitter locations and the power coverage, while the encoder expands feature maps from the encoder into a full-resolution power coverage map. 
We also propose improvement to the basic UNET by replacing the pooling layers with strided convolutions and combining the UNET with inception modules as introduced by GoogleNet \citep{szegedy2015going}. 
These modules allow increasing the network depth and width while keeping the computational cost constant \citep{zhao2017survey}. 
To scale up the model to the higher resolution instances, we consider architectures where we gradually increase the number of convolution layers and the complexity of inception modules. 
More convolutional layers produce more feature maps to learn from which is a logical approach when resolution increases. 
On the other hand, inception modules enable learning large and small scale propagation patterns in parallel when moving from low building density areas to the higher density areas.

Levie et al.~\citep{levie2021radiounet} first proposed to use UNET to estimate path loss function in radio propagation. 
In their implementations, the authors use the RadioMapSeer dataset with 700 simulated radio maps and 80 transmitter locations per map. 
They use three propagation models: Dominant Path Model (DPM) that considers only the path with the highest energy contribution, intelligent ray tracing - 2 (IRT2) that considers two reflections and intelligent ray tracing - 4 (IRT4) with four reflections. 
Maps are taken from OpenStreetmap. 
Levie et al.~\citep{levie2021radiounet} also applied RadioUNET to demonstrate that radio signal estimation can be used for coverage classification and pathloss fingerprint based localization. 


The contributions of our study can be summarized as follows:
\begin{itemize}
    \item We consider several CNN and UNET variants, and compare those against the recently developed RadioUNET model, which is a UNET-based deep learning architecture.
    Different than the RadioUNET model, our UNET models include enhancements such as strided convolutions and inception layers, which were previously shown to be effective in improving prediction performance for various computer vision tasks.
    
    
    \item We consider datasets with different characteristics in our analysis. 
    Our primary dataset is obtained from five urban regions using Wireless Insite ray tracing simulation software, which allows 6 reflections, and generate complex propagation patterns.
    We utilize a full 3D representation of buildings that ray tracing uses to determine the propagation patterns. We then define a 2D radio map at a given elevation as input to our model.
    As Wireless Insite is a propriety software, data generation is expensive.
    Accordingly, we consider various data augmentation strategies to generate sufficiently large datasets for deep learning model training.
    We also experiment with the large datasets provided by Levie et al.~\citep{levie2021radiounet}.
    
    \item We perform a detailed numerical study to highlight the power of CNN-based models for the radio signal power prediction task.
    We experiment with different frame sizes e.g., from $32\times 32$ frames to $256\times 256$ frames, and empirically show how increase in model complexity helps prediction performance for larger frames.
    We also examine the practical use cases for prediction models including power estimation for multiple transmitters, and capturing the impact of moving objects in real time.
    
\end{itemize}

The rest of this paper is organized as follows. Section \ref{sec:background} reviews the recent literature on deep learning models including, UNET and RadioUNET models, that are used for similar prediction tasks. The dataset structure, feature engineering methods and models' structures are discussed in Section \ref{sec:Methodology}. The results from a detailed numerical study are reported in Section \ref{sec:Results}. Finally, a summary of our work and future research directions are provided in Section \ref{sec:conclusions}.

\section{Background}\label{sec:background}
In this section, we briefly review the recent relevant studies on predictive modeling in coverage prediction. 

Earlier studies on coverage/propagation prediction focus on empirical and deterministic models 
\citep{erceg1999empirically, hata1980empirical}. 
Empirical models incorporate sets of measurements from a sample environment to model the propagation characteristics. 
However, they fall short of generalizing the predictions to new environments. 
Moreover, physical laws are used for deterministic modeling of new environments, but it suffers from expensive computational complexity. 
On the other hand, many recent studies incorporated different machine learning algorithms such as Neural Networks (NN), Random Forest (RF) and Support Vector Machines (SVM) for propagation prediction and outperformed the deterministic and empirical models \citep{angeles2015neural}. 
Such simple machine learning methods can be used for less complex problems to avoid overfitting and high computational complexity. 
In addition, they are shown to be effective when there is a limited available data.
For instance, Mohammadjafari et al.~\citep{mohammadjafari2020machine} showed that generic machine learning methods such as Generalized Linear Model (GLM) with an extended feature engineering process achieves a comparable performance to NN models in radio signal prediction for a small dataset.

Complex deep learning models such as CNNs and the variants have recently become popular for power coverage prediction. 
CNN is a deep learning architecture that is constructed by a set of layers. The core building blocks of a CNN are the convolutional layers that convolve the input with a kernel, then apply an activation function, typically a rectified linear unit (ReLU) or Sigmoid function. CNN design consists of defining the number and sequence of layers and the size of kernels. CNN has wide applications ranging from image classification \citep{simonyan2014very, szegedy2015going}, object detection \citep{girshick2015region, he2015spatial, he2016deep} and image retrieval systems \citep{radenovic2016cnn, radenovic2018fine}. 
Semantic image segmentation is one of the applications where CNNs were applied successfully \citep{ciresan2012deep, gupta2014learning, pinheiro2014recurrent}. Semantic image segmentation is a task in which we label specific regions of an image according to what object is being shown. This task is commonly referred to as dense prediction \citep{long2015fully} as it is the progression from coarse to fine inference that makes a prediction at every pixel.

When used for classification, the last layers of a CNN consist of one or more fully connected layers that predict the image label. Fully convolutional network (FCN) was introduced by Long et al.~\citep{long2015fully} and is a special CNN where the final fully connected layers are replaced with convolutional layers. In this way, FCN can be trained end-to-end, pixels-to-pixels, which is very suitable for the task of semantic segmentation. Long et al.~\citep{long2015fully} proposed well-studied classification networks as encoder along with a decoder module with transposed convolutional layers to upsample the coarse feature maps into a full-resolution segmentation map. However, it was challenging to produce fine-grained segmentation from the low resolution encoder output. The authors mitigated this issue by upsampling the encoded representation in stages and by adding skip connections from earlier layers. Skip connections provide the necessary higher resolution details to reconstruct accurate segmentation boundaries. 

Ronneberger et al.~\citep{ronn2015} propose an enhanced version of the FCN called the UNET. While the encoder-decoder structure is maintained in the UNET, these components are modified to be symmetrical and contain the same number of layers. The downsampling layers in the encoder have corresponding upsampling layers in the decoder. Moreover, in the UNET, skip-connections directly connect opposing contracting and expanding convolutional layers to provide more detail in the segmentation result. The UNET architecture was used in medical images segmentation \citep{litjens2017survey,milletari2016v}, super resolution \citep{lim2017enhanced} and image-to-image translation \citep{yi2017dualgan}. The original UNET implementation was extended to 3D by Cicek et al.~\citep{cciccek20163d}, and extended to other variants by Drozdzal et al.~\citep{drozdzal2016importance} and Jegou et al.~\citep{jegou2017one}. In a recent study, Levie et al.~\citep{levie2021radiounet} proposed UNET architectures to estimate path loss function in radio signal propagation.

\section{Methodology}\label{sec:Methodology}
In this section, we first describe our single region dataset and the preprocessing steps undertaken to obtain a multi-region dataset. Then, we present the models and their different components. Finally, we specify the details of the model training such as hyperparameters and the choice of evaluation metric.

\subsection{Dataset}
\begin{figure}[!ht]
    \centering
    \includegraphics[width=0.39\columnwidth]{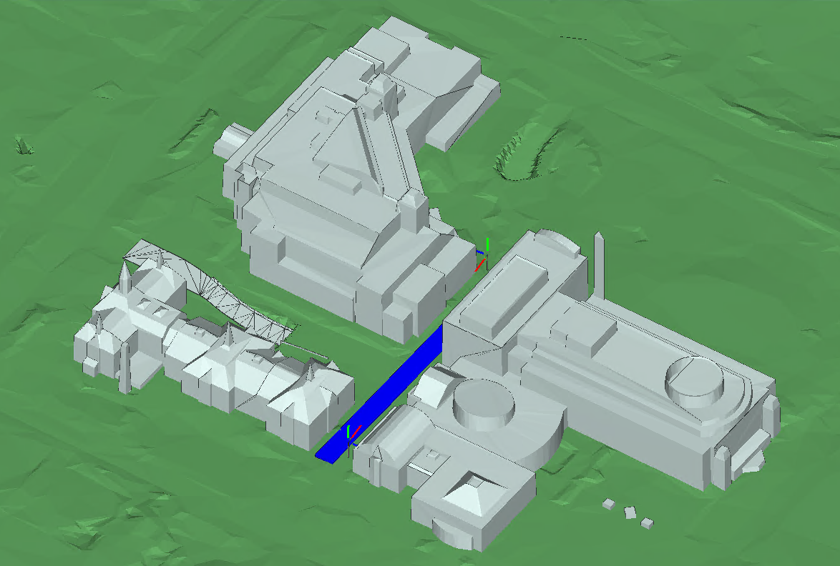}
    \caption{A highly detailed 3D surface of Ottawa City Hall}
    \label{fig:ottawa_model}
\end{figure}

Our \textit{fixed-region} datasets consist of a set of simulated radio propagation scenarios from five different non-overlapping areas in downtown Ottawa, generated using a Wireless Insite ray-tracing software \citep{insite}, which we refer to as Wireless Insite dataset.
We chose to use detailed 3D maps of downtown Ottawa, a sample of which is shown in Figure \ref{fig:ottawa_model} to run the 3D ray-tracing simulations. 
More specifically, the dataset contains various objects with rough surfaces that interacts with the rays in complex ways. Moreover, there can be gaps in the objects such as bridges, and rays at different levels can pass or reflect from such objects. The detailed representations of the objects and six ray interactions result in a complex power distribution that is representative of a real-life measurement.

\begin{figure}[!ht]
    \centering
    \subfloat[Region map ]{\includegraphics[width=0.27\columnwidth]{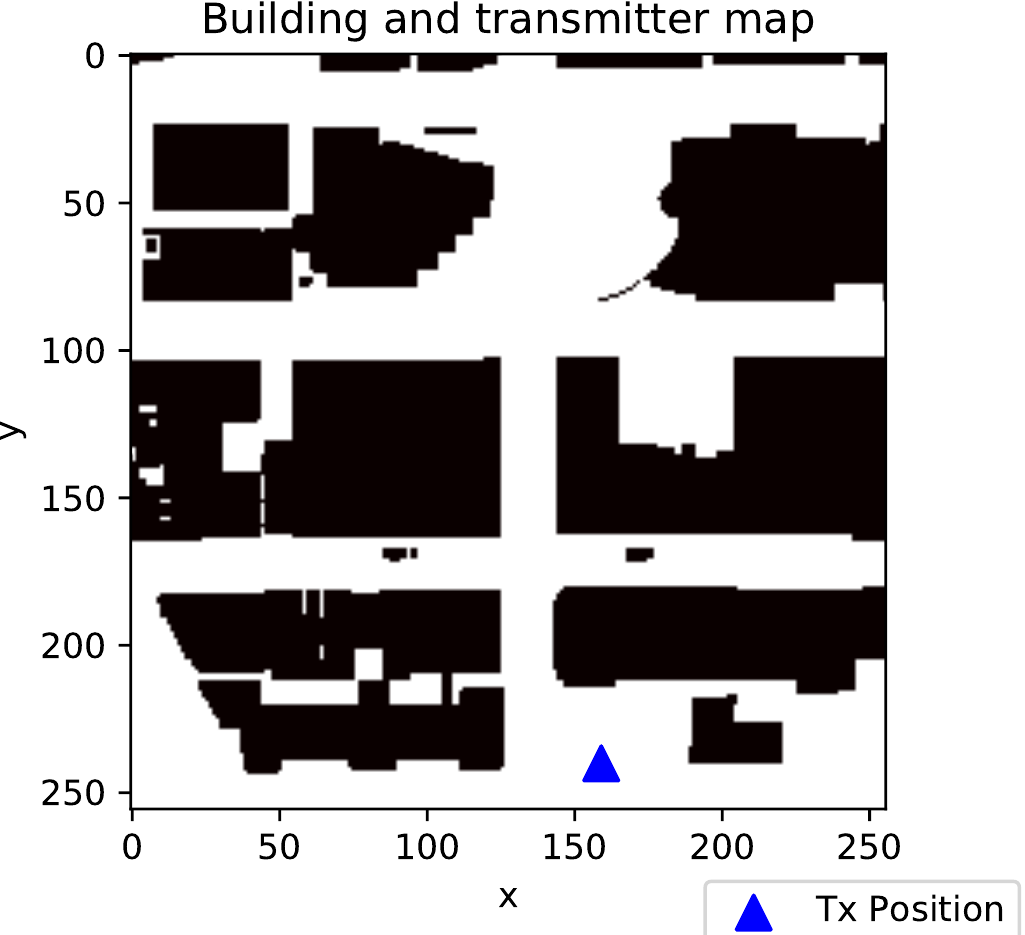}
    \label{fig:buildigns}
    }
    \subfloat[Coverage]{\includegraphics[width=0.27\columnwidth]{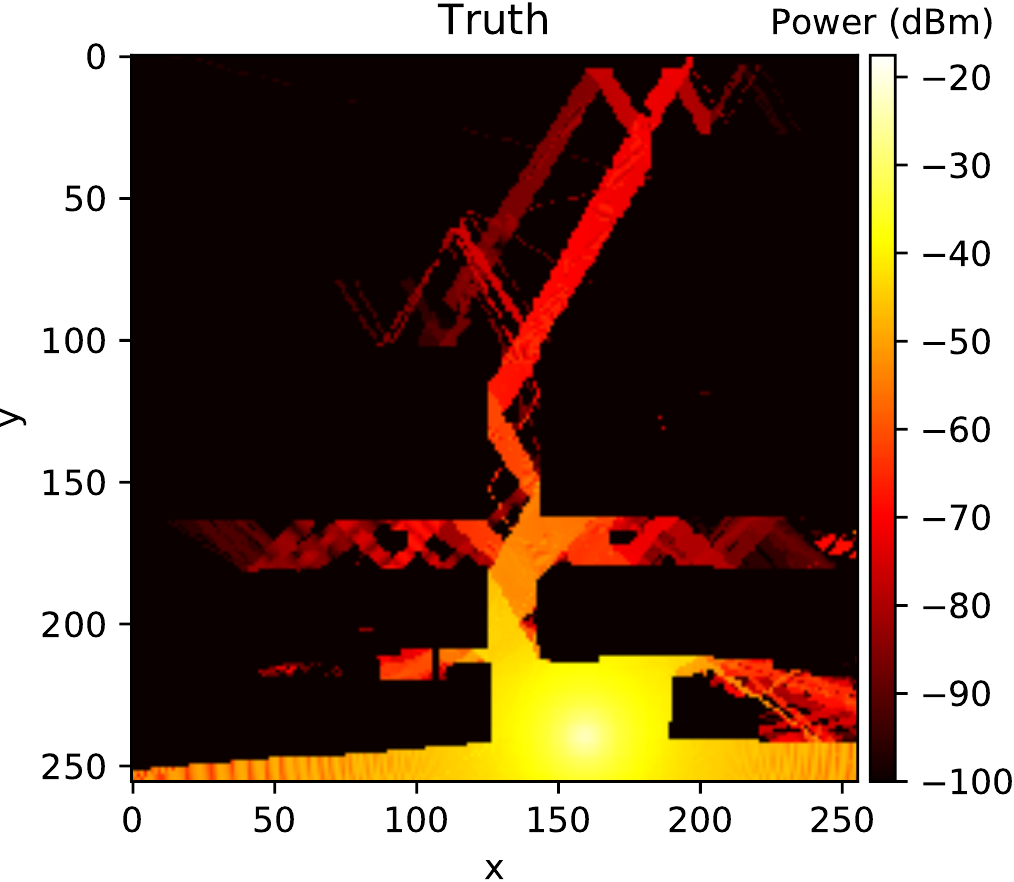}
    \label{fig:sample}
    }
    \caption{$256\times 256$ dataset sample. Black pixels in figure (a) represents the building locations. }
    \label{fig:dataset_sample}
\end{figure}

Each of the five areas has 1,000 power coverage simulations for random transmitter locations and has the size of 257$\times$257 pixels. 
The transmitter is assumed to use an isotropic antenna, producing omnidirectional radio waves. 
The height and power of the antenna were set at maximum to 6 metres and 50 Watts respectively. 
The propagation model in ray tracing considers 6 reflections on 3D buildings which results in more complex patterns than with DPM or IRT4. 
We then consider a 2D plane at a given elevation where each pixel presents an area of 1$\times$1 metres defined by its Cartesian coordinates $x$ and $y$ and a power measurement in dBm. 
A sample of power coverage is presented in Fig.~\ref{fig:dataset_sample}. 
Our deep learning models require input that is divisible by two. 
Thus, we cropped one pixel from the corners to obtain $256\times256$ samples. 
The samples where the transmitter is removed after cropping are excluded from the dataset.

\subsubsection{Constructing multi-region dataset}
As we aim to solve the power coverage estimation problem for urban areas, we are primarily interested in training the prediction model on different urban environments. 
However, the fixed-region dataset contains only five different environments. 
Thus, we selected smaller frames from the fixed-region dataset to create new multi-region datasets of sizes $32 \times 32$, $64 \times 64$, $128 \times 128$, and $256 \times 256$. 

We summarize the procedure to generate $32 \times 32$ instances as follows. 
First, for every $256 \times 256$ sample in a fixed-region dataset, a $32 \times 32$ sliding window is moved through the sample with a stride of $3$. 
We only consider frames that include the transmitter and building blocks since we aim to estimate power coverage of the transmitter in an urban environment. 
Some frames contain a high rate of reflections coming from buildings outside the frame itself.
These are problematic cases since during training, the model only receives the 32$\times$32 frame and it cannot be informed about the reflections coming from outside the frame. 
Fortunately, most of these cases happen when the transmitter is near the edges of the frame. 
Thus, we applied a padding of 5 pixels from all sides and all frames where the transmitter is near the edges are removed from the multi-region dataset. 
Similar data transformation and augmentation approaches are used for creating other frame sizes as well.

\begin{figure}[!ht]
    \centering
    \includegraphics[width=0.26\columnwidth]{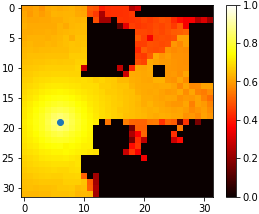}
    \caption{A sample $32\times 32$ from the dataset. The blue dot shows the transmitter location.}
    \label{fig:sample_scale}
\end{figure}

After the samples are generated, we apply a minimum threshold of $-100~ \rm{dBm}$, since the coverage estimations are not deemed to be interesting below this value. 
We apply min-max normalization over the power coverage, and scale the samples between $0-1$. 
A sample $32\times 32$ frame from the generated dataset is shown in Fig.~\ref{fig:sample_scale}.

The dataset is divided into train and test sets based on the frames' original location on the fixed-region map. 
We consider the upper-left corner of each frame as the reference point. 
The training set contains all frames whose upper-left corner $x$-coordinate is below 60. 
Test set frames are above the $x$-coordinate of 80. 
We leave a gap of 20 between the train and test set regions to prevent overlapping. 
Since the goal is to obtain a generalizable model, train and test sets contain samples from different areas. 
The total number of data instances are 508k, 374k, 334k and 20k for $32 \times 32$, $64 \times 64$, $128 \times 128$, and $256 \times 256$ frame sizes respectively.


\subsubsection{RadioMapSeer dataset}
Levie et al.~\citep{levie2021radiounet} provided 56,000 simulated radio maps in 700 different city locations and 80 transmitter location maps over a 2D $256\times 256$ $m^2$ grid, which we refer to as RadioMapSeer dataset\footnote{\url{https://radiomapseer.github.io/}}. 
The simulations are generated using DPM \citep{wahl2005dominant} and IRT2 \citep{rautiainen2002verifying}.
While we mainly consider Wireless Insite datasets in our analysis, we also experiment with RadioMapSeer dataset to compare our results against Levie et al.~\citep{levie2021radiounet}'s findings.

\subsubsection{Feature engineering}
The dataset consists of two sets of input features: transmitter location and urban environment. 
We tested four different approaches to feed these features to the models. 
In the first approach, both the building location and the environment are provided in the same input channel. 
Consequently, the input is a 2D image, where the pixel values are set to one and two for building and transmitter locations, respectively. 
In the second scenario, we divide the input into two separate channels. 
Both channels are binary 2D images. 
The first encodes building locations with ones (i.e., a white colour) while in the second, a one corresponds to the transmitter location. 
In the third and fourth scenarios, the urban environment is represented by the same 2D image, but in this case the transmitter is represented by a 2D image where the value of each pixel is equal to its distance from the transmitter. 
We used Euclidean distance and inverse square distance as the distance metrics in the third and fourth scenarios, respectively. 
In all cases, the default pixel value is set to zero.

We performed preliminary analysis with all four input scenarios and observed that the second scenario provides the best performance in terms of the Mean Absolute Error (MAE). 
Thus, we retain these input features in our numerical experiments.




\subsection{Deep learning models}
We train two types of CNN models for the power coverage prediction, namely, vanilla CNN and UNET. 
Below we provide a brief description of these models along with various model enhancements such as strided convolutions and inception layers.
\subsubsection{Baseline CNN}
\begin{figure*}[!ht]
    \centering
    \includegraphics[width=0.8\textwidth]{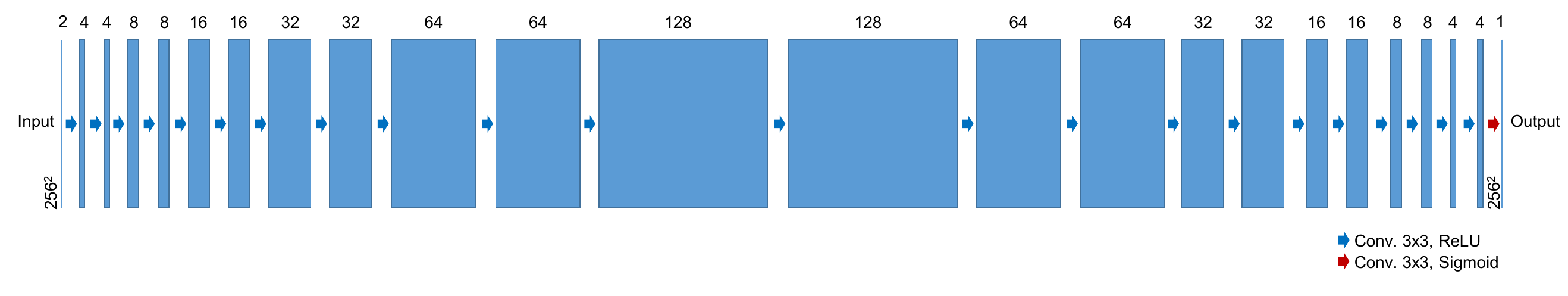}
    \caption{Baseline CNN model architecture}
    \label{fig:cnn_arch}
\end{figure*}

We consider a vanilla CNN model as a baseline prediction model. 
The model consists of 24 2D convolutional layers with $3\times3$ kernels. Each layer has 32 filters, ReLU activations and identical padding so that the output size of the layer is same as the input size. Fig.~\ref{fig:cnn_arch} demonstrates the adopted CNN architecture.

\subsubsection{RadioUNET}
RadioUNET architecture was proposed by Levie et al.~\citep{levie2021radiounet}, which consists of 41 convolutional layers.
The detailed structure including dimension of the images in the layers, channels and kernel sizes is presented in Table~\ref{tbl:RadioUnet}.
\setlength{\tabcolsep}{4pt}
\renewcommand{\arraystretch}{1.3}
\begin{table*}[!ht]
\begin{small}
\caption{RadioUNET structure}
\label{tbl:RadioUnet}
\begin{center}
\begin{tabular}{|c|c|c|c|c|c|c|c|c|c|c|c|c|c|c|c|c|c|c|}
\hline 
 & \textbf{In} & \textbf{L1} & \textbf{L2} & \textbf{L3} & \textbf{L4} & \textbf{L5} & \textbf{L6} & \textbf{L7} & \textbf{L8} & \textbf{L9} & \textbf{L10} & \textbf{L11} & \textbf{L12} & \textbf{L13} & \textbf{L14} & \textbf{L15} & \textbf{L16} & \textbf{out}\\
\hline
\textbf{Resolution}  & 256 & 256 &  128 & 64 & 64 & 32 & 32 & 16 & 8 & 16 & 32 & 32 & 64 & 64 & 128 & 256 & 256 & 256 \\
\hline
\textbf{Channel}  &2/3 & 6 & 40 & 60 & 80 & 100 & 120 & 200 & 400 & 400 & 240 & 200 & 160 & 120 & 80 & 29 & 32 & 1 \\
\hline
\textbf{Filter} &  3 & 5 &  5 & 5 & 5 & 3 & 5&5 & 4& 4& 3&6 &5 & 6& 6 & 5 & 2& -\\
\hline
\end{tabular}
\end{center}
\end{small}
\end{table*}

\subsubsection{UNET with Strided Convolutions and Inception}
\begin{figure*}[!ht]
    \centering
    \includegraphics[width=0.85\textwidth]{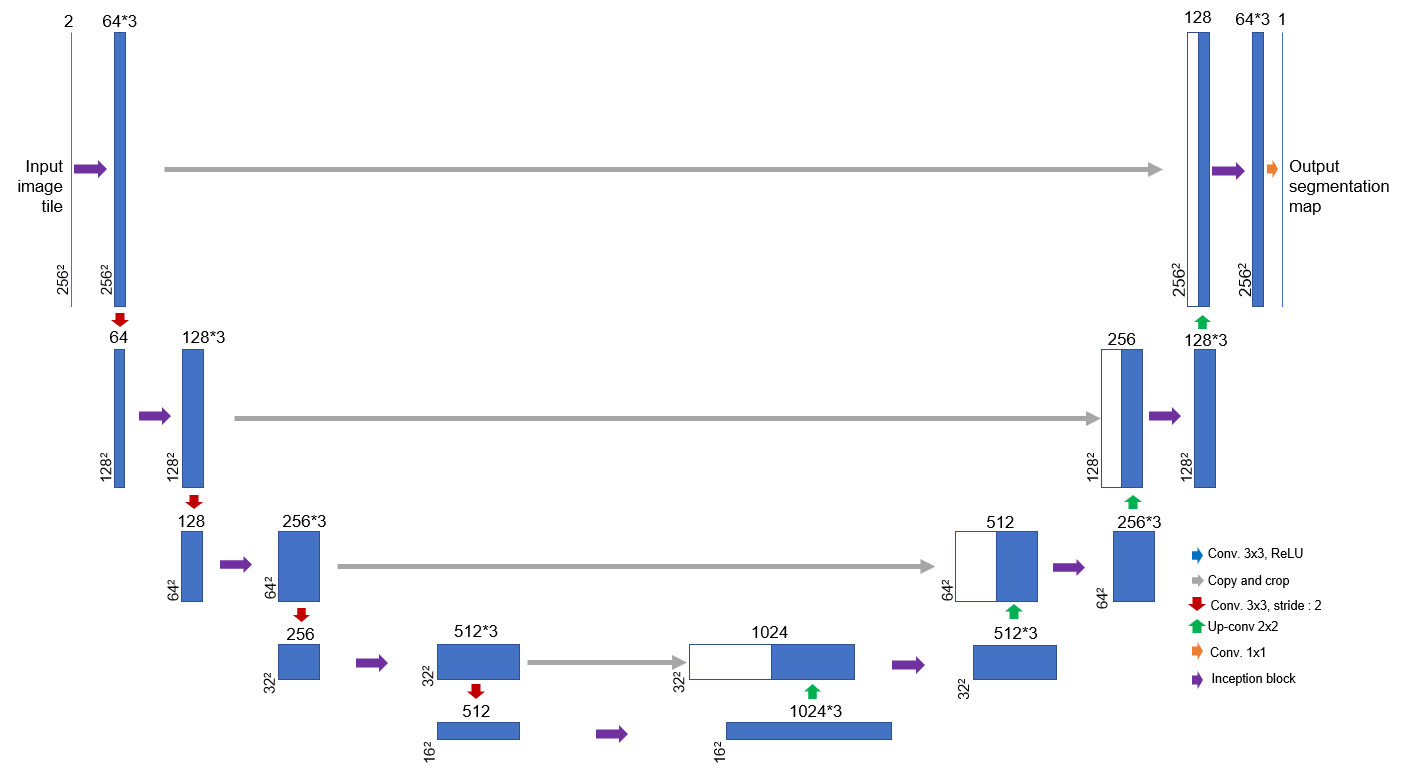}
    \caption{UNET model structure}
    \label{fig:unet_arch}
\end{figure*}

UNET is a special type of CNN which was originally developed for biomedical image segmentation~\citep{ronn2015}. 
The encoder-decoder architecture of the UNET is illustrated in Fig.~\ref{fig:unet_arch}. 
The left side of the U shape is called the encoder. 
It mainly consists of $3\times3$ convolutions, each followed by a ReLU and $2\times2$ max pooling layers with a stride of two for downsampling. 
The number of features are doubled at each downsampling step.

The decoder to the right hand side of the network uses transposed convolutions for upsampling to increase the output size to the same size as the input. 
The decoder consists of $2\times2$ convolutions which halves the number of feature channels, a concatenation with the corresponding encoder layer's output, and two $3\times3$ convolutions, each followed by a ReLU activation. 
The last layer is a $1\times1$ convolution with a single channel. 
The network depicted in Fig.~\ref{fig:unet_arch} has 24 convolutional layers in total.

CNNs with pooling layers aggregate the nearby pixels, therefore the spatial information is lost. 
UNET solves this issue by an encoder-decoder architecture and skip connections. 
Each encoder layer has a skip connection with its corresponding decoder layer to allow the decoder network to retrieve fine-grained details from the encoder.

The performances of the CNN (and UNET) models can be enhanced by incorporating special layers such as inception and strided convolutions that are customized for the power coverage prediction task. 
Below we briefly describe how these model enhancements are incorporated into the UNET architecture.

\paragraph{Strided convolutions.}
The basic UNET model uses max pooling layers to downsample images. However, Springenberg et al.~\citep{springenberg2014striving} showed that replacing max pooling layers with strided convolutions can improve the accuracy of the model with the same depth and width. 
This improvement comes with a cost as max pooling is a fixed operation in which the model does not learn any new parameters, making it computationally cheaper. 
On the other hand, the same procedure with strided convolutions requires the model to learn new parameters and thus increases the computational cost.

\paragraph{Inception layers.}
Inception was first introduced in GoogleNet \citep{szegedy2015going}, which was considered to be a milestone in CNN classification problems. 
Before inception, most classifiers stacked deeper convolution layers to increase performance, which is computationally expensive, and makes the model prone to overfitting.
The premise behind inception idea is that salient parts in an image can have large variation in size. 
For example, to detect the presence of a cat in an image, the animal head, crucial in classification, can have different sizes, thus the right kernel size for convolutions is hard to choose. 
Larger kernel sizes are preferred if information covers a large area in the image, while smaller kernels allow for a better local object detection. 
In Fig.~\ref{fig:sample}, we can observe global propagation patterns in the upper part of the image and smaller pattern in the bottom part where the density of buildings is higher. 
Using inception in this context is helpful as multiple kernels can lead to a better learning of both large and small scale reflection patterns by combining the output of different kernels.

\begin{figure*}[!ht]
    \centering
    \includegraphics[width=0.67\textwidth]{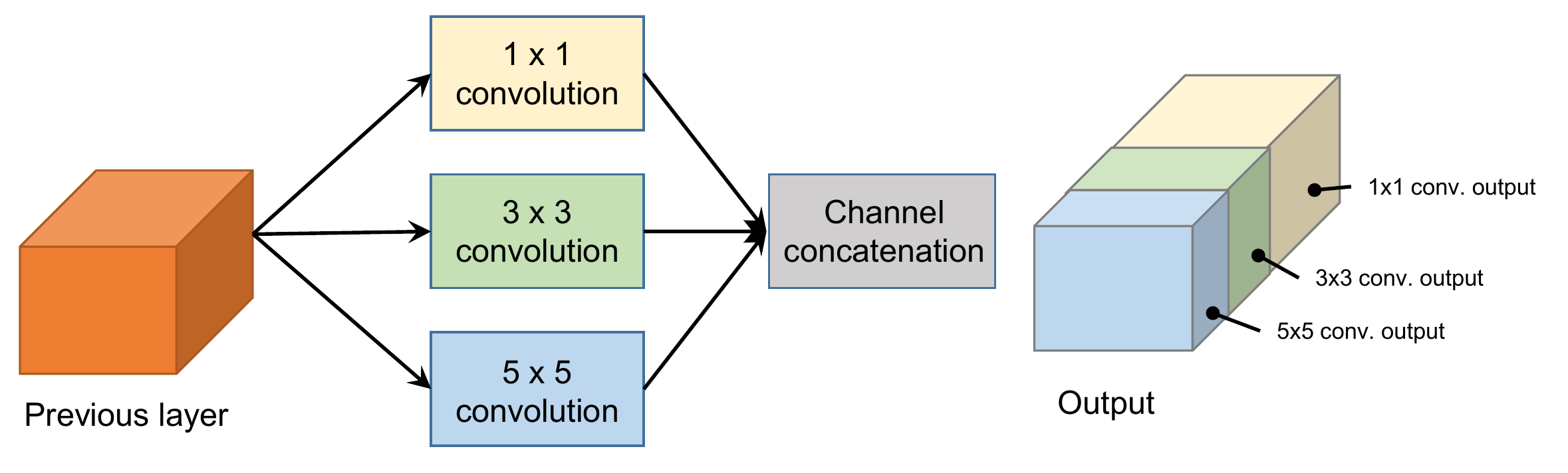}
    \caption{Inception module structure: an inception module combines the output of multiple kernels. This favours detecting global and local patterns. The number of higher size filters is generally smaller to reduce computational cost.}
    \label{fig:Inception_structure}
\end{figure*}

In this work, we use three parallel kernel sizes in the inception structure depicted in Fig.~\ref{fig:Inception_structure}. 
Specifically, three different kernel sizes are implemented on the input to extract features and combine them on the channel dimension. 
This means the output has the same dimension as the input and a channel dimension three times the input channel dimension. 
The output of the previous layer passes through $1\times1, 3\times3$ and $5\times5$ kernels. 
The outputs are concatenated by the channel dimension before being passed to the next layer. 
The kernel size of $1\times1$ is computationally cheaper compared to other kernel sizes and is typically employed to change the number of channels.

\begin{figure*}[!ht]
    \centering
    \includegraphics[width=0.6\textwidth]{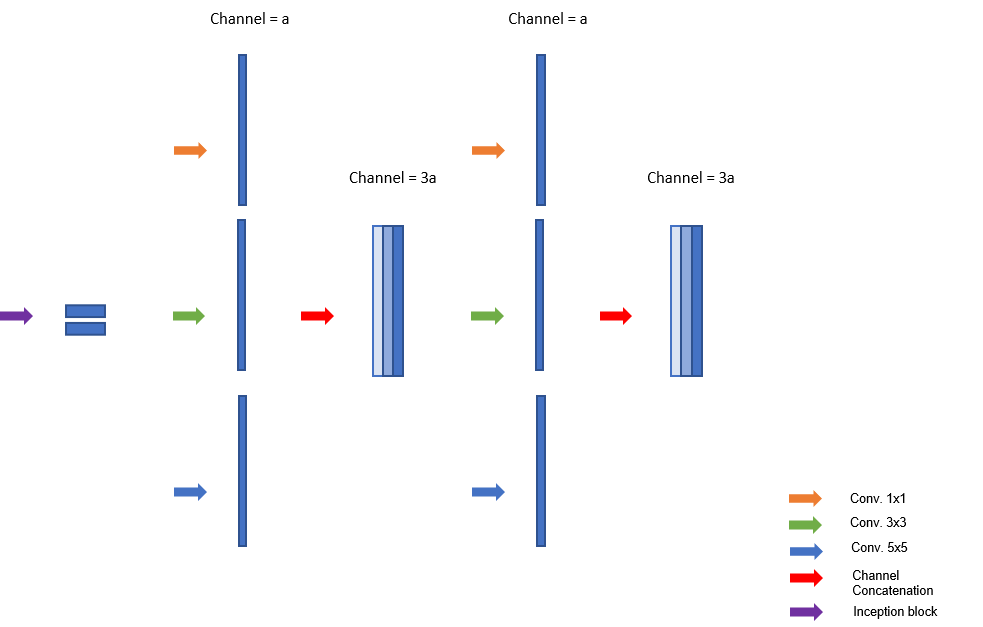}
    \caption{A sample two-layered inception structure (UNET-SI-65).}
    \label{fig:Inception_examples}
\end{figure*}

The purple arrows in Fig.~\ref{fig:unet_arch} indicate the proposed inception blocks designed to increase the complexity of the models. 
We considered four different UNET configurations in our analysis: 
UNET-SI-37, UNET-SI-65, UNET-SI-73 and UNET-SI-91. 
UNET-SI-37 contains 37 convolutional layers in total and 13 convolutional layers in the inception structures.
The UNET-SI-65 model uses an inception structure (see Fig.~\ref{fig:Inception_examples}) that passes the output of the first inception layer through three different kernel sizes for the second time and merges those at the end. 
This model contains 65 2D convolutional layers and 23 convolutional layers in the inception structure. 
The UNET-SI-91 uses an inception structure with a depth that is 3 times the original UNET-SI-37 model. 
Because of its higher complexity, this model could not be applied to $256\times256$ frames with the hardware used in the experiments. 
As a solution, we proposed an architecture of intermediate complexity denoted UNET-SI 73 that has the same inception structure as the UNET-SI-65 but with two strided convolutional 2D layers instead of one.


\subsection{Model training}
We perform supervised learning on the multi-region dataset of urban environments to estimate the power coverage at each location based on a given transmitter location and environment. 
The MAE is used as the accuracy metric on the normalized power values. 
After training, MAE scores are denormalized and reported as they are easier to interpret. 
The MAE shows the error average in dB in a given frame. 
We performed hyperparameter tuning for CNN, RadioUNET and UNET variants by considering a range of values over batch size, kernel size and learning rate.
We reduced the batch size from 128 to 8 as the frame sizes increase from $32\times32$ to $256\times256$, and increased the number of epochs from 40 to 200 for more complex models.
Early stopping with a patience of three is used to monitor the test loss. 
In the final configurations, the Adam optimizer~\citep{adam} is used with an initial learning rate of 0.001 and exponential decay rates of 0.9 and 0.999 for $\beta_1$ and $\beta_2$, respectively, for all the models. 
The summary statistics regarding the models considered in our analysis is provided in Table~\ref{tbl:Models}.
\setlength{\tabcolsep}{4pt}
\renewcommand{\arraystretch}{1.3}
\begin{table*}[!ht]
\begin{small}
\caption{{CNN and UNET model structures 
}}
\label{tbl:Models}
\begin{center}
\scalebox{0.73}{
\begin{tabular}{|c|c|c|c|c|}
\hline 
\textbf{Model} & \textbf{Inception Structure} & \textbf{\# of parameters} & \textbf{\# of convolution layers} & \textbf{\# of Convolution layers in the inception structure}\\
\hline
UNET-SI 37  & a & 90M& 37 &13 \\
\hline
UNET-SI 65  & b & 274M& 65 &23 \\
\hline
UNET-SI 91 & c & 457M& 91 & 31 \\
\hline
UNET-SI 73 & b &278M& 73 & 23 \\
\hline
RadioUNET & - & 4M & 41 & -\\
\hline
\end{tabular}
}
\end{center}
\end{small}
\end{table*}

\section{Results}\label{sec:Results}
We performed detailed numerical experiments to demonstrate the effectiveness of UNET models for the power coverage prediction task. 
We start with a performance comparison between different deep learning architectures (e.g., CNNs and UNET) using $32\times 32$ frames.
We next examine the impact of various settings such as kernel size and strided convolutions.
Then, we assess the performance of the coverage prediction models over larger frame sizes.
Lastly, we discussed how the UNET models can be used in practice. 
All models were implemented using a NVIDIA TESLA P40 GPU with 24 GB of GPU RAM.

\subsection{CNNs vs UNET}
The performance of the UNET and the CNN models with different kernel sizes and for five repeats are reported in Table~\ref{tbl:Baseline models}, which shows model parameters, MAE and training run time. 
For this analysis, we only considered $32\times 32$ frames for the sake of computational efficiency as larger frame sizes typically lead to longer training times.
Kernel size of 5 provides the best performance for both CNN and UNET models. 
The best overall performance is obtained by UNET model with a kernel size of 5, which leads to 5.2\% improvement in terms of average MAE compared to its CNN counterpart (2.69 vs 2.55), though it has significantly longer run times (0.52 hours vs 2.4 hours). 
We note that benefits of UNET is amplified for specific instances, e.g., MAE improvements are 13.6\% (2.64 vs 2.28) and 11.6\% (2.77 vs 2.45) over min and max MAE values. 
Additionally, we observe that UNET models involve a significantly larger amount of model parameters and, on average, the UNET model run times are approximately two to three times greater than the CNN models. 
We perform the rest of the numerical experiments using the UNET models as they provide better overall predictions.

\setlength{\tabcolsep}{12pt}
\renewcommand{\arraystretch}{1.2}
\begin{table*}[!ht]
\caption{{The baseline CNN and UNET model performances reported for 5 repeats over $32\times 32$ frames.}}
\centering
\scalebox{0.8}{
\begin{tabular}{ccccccc}
\toprule
\multirow{2}{*}{Model} &  \multirow{2}{*}{Kernel size} & \multicolumn{3}{c}{MAE (dB)} & \multirow{2}{*}{Time (hr)} & \multirow{2}{*}{\# of parameters}\\
 \cline{3-5}
 &  &  min & max & average & & \\
\toprule
\multicolumn{1}{c}{CNN}   &       3     &  2.67 & 2.73 & 2.71   & \multicolumn{1}{c}{0.88}        & \multicolumn{1}{c}{204.35K}              \\ \midrule
\multicolumn{1}{c}{CNN}   &    5       &  2.64 &  2.77 & 2.69    & \multicolumn{1}{c}{0.52}        & \multicolumn{1}{c}{566.33K}                \\ \midrule
\multicolumn{1}{c}{CNN}   &   7        & 2.82 & 35.71 &  13.84  &  \multicolumn{1}{c}{0.51}        & \multicolumn{1}{c}{1.10M}               \\ \midrule
\midrule
\multicolumn{1}{c}{UNET}  & 3           & 2.25  & 3.92 & 2.97   & \multicolumn{1}{c}{1.13}        & \multicolumn{1}{c}{31.03M}               \\ \midrule
\multicolumn{1}{c}{\textbf{UNET}}  & \textbf{5}           & \textbf{2.28}  & \textbf{2.45} & \textbf{2.55}   & \multicolumn{1}{c}{\textbf{2.40}}        & \multicolumn{1}{c}{\textbf{81.23M}}               \\ \midrule
\multicolumn{1}{c}{UNET}  & 7           & 2.36  & 4.17 & 3.25   & \multicolumn{1}{c}{4.31}        & \multicolumn{1}{c}{156.54M}              \\ \midrule
\end{tabular}
}
\label{tbl:Baseline models}
\end{table*}

\subsection{The impact of kernel size for UNET performance}
\begin{figure*}[!ht]
    \centering
        
        \subfloat[Region map ]{\includegraphics[width=0.25\textwidth]{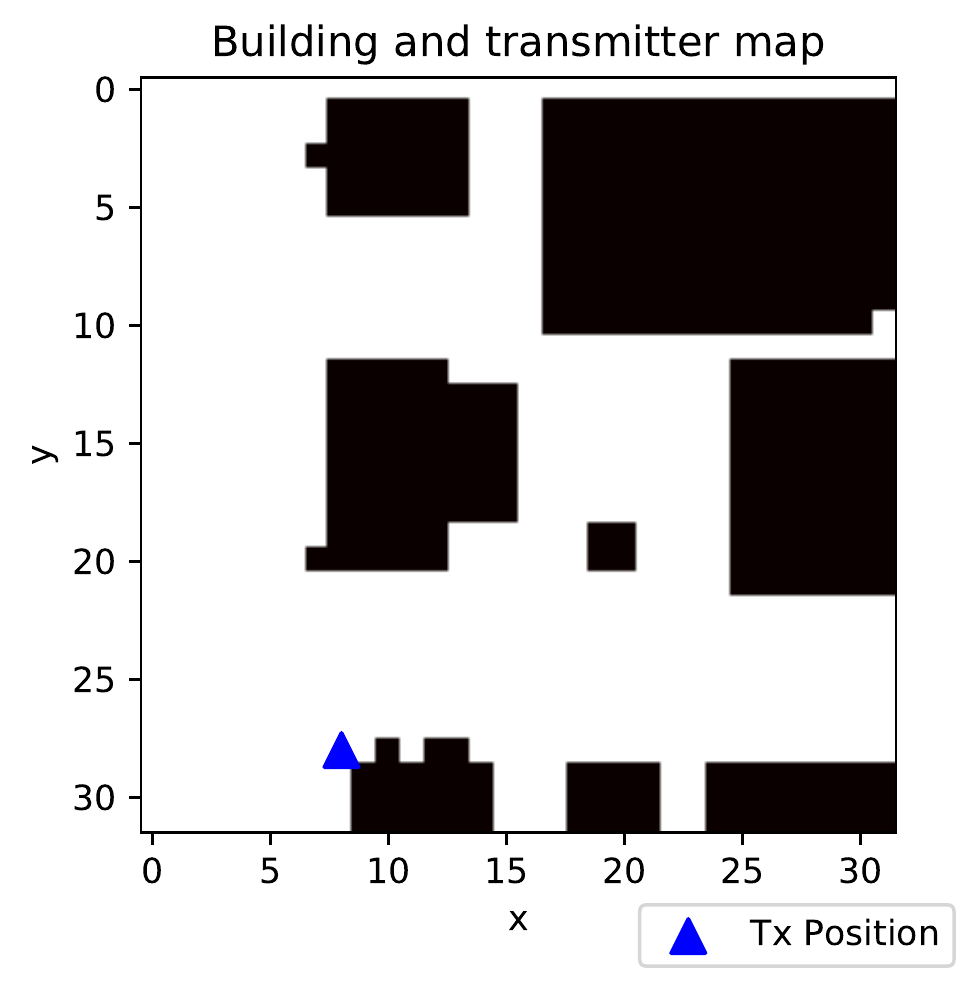}
        \label{fig:Building_20000}
        }
        \qquad
        \subfloat[UNET model with Kernel 3$\times$3 prediction\label{fig:Kernel3_20000 prediction} ]{\includegraphics[width=0.75\textwidth]{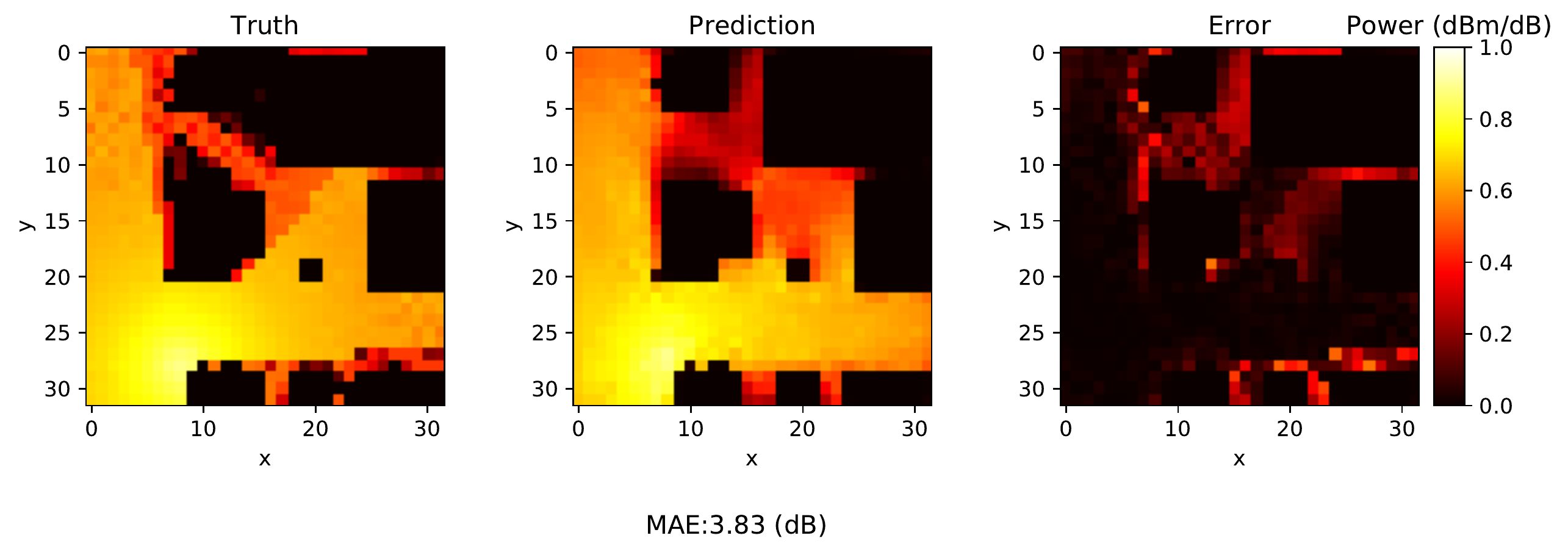}
        }
        \qquad
         \subfloat[UNET model with Kernel 5$\times$5 prediction]{\includegraphics[width=0.75\textwidth]{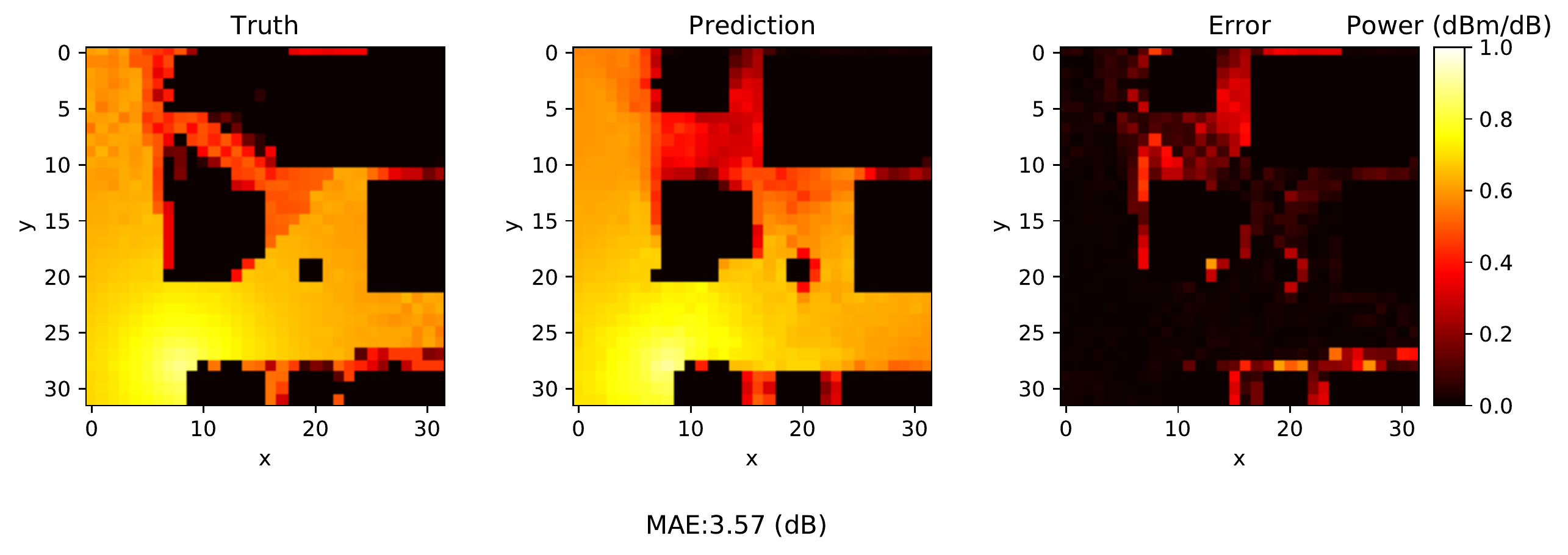}
        \label{fig:Kernel5_20000 prediction}
        }
        \qquad
         \subfloat[UNET model with Kernel 7$\times$7 prediction\label{fig:kernel7_20000 prediction}]{\includegraphics[width=0.75\textwidth]{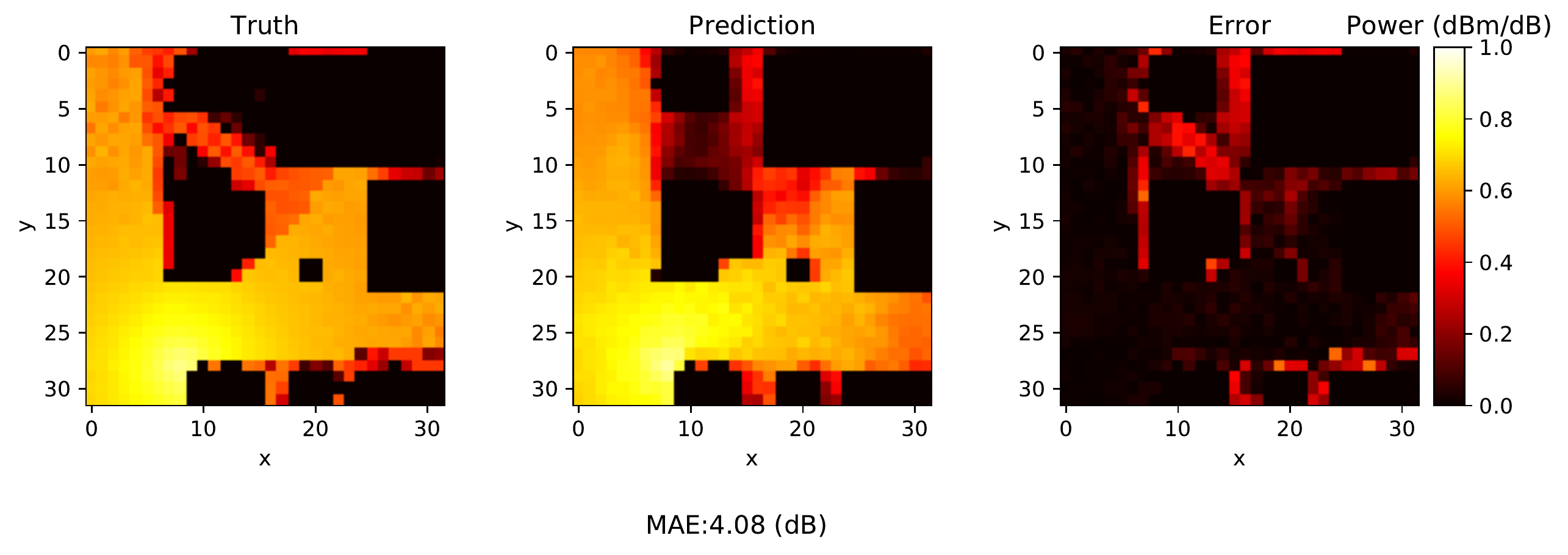}
        }
    \caption{UNET predictions for different kernel sizes ($32\times 32$ frames)}
    
    \label{fig:predictions_kernel_variations}
\end{figure*}

We next provide visual demonstrations of model predictions for different kernel sizes over a sample region map represented by $32\times 32$ frames. 
Fig.~\ref{fig:predictions_kernel_variations} illustrates the UNET model predictions with the same sample and with three different kernels. We oberve that kernel size of 5$\times$5 performs best with an MAE of 3.57 dB while a lower kernel of 3$\times$3 and a higher kernel of 7$\times$7 yield less accurate predictions. For all cases, the power coverage is close to the ground truth around the transmitter located at coordinate (8,28). All three models are able to capture the effect of buildings on decreasing the incoming power by casting a radio shadow behind buildings. However, the difference in accuracy can be clearly observed near the building block located at coordinate $(20,20)$. Given the small dimension of the building block and transmitter location, the ground truth shows no significant power attenuation around this block. The prediction with kernel 5$\times$5 is the closest to the truth which suggests it is the optimal kernel. A smaller kernel of 3$\times$3 considers only close pixels, this may explain the fuzziness observed around the small building block in Fig.~\ref{fig:Kernel3_20000 prediction}. A higher kernel of 7$\times$7 considers a much broader area around every pixel and this produces power distribution patterns that are not consistent with the ground truth in Fig.~\ref{fig:kernel7_20000 prediction}.

\subsection{Performance of the enhanced UNET}
We trained enhanced versions of the UNET that include inception modules and strided convolutions. 
The results obtained with the different variants over $32\times 32$ frames are reported in Table~\ref{tbl:UNET Variations}. 
Replacing max-pooling layers with strided convolutions and a kernel size of 3 decreased the average MAE from 2.97dB to 2.39dB (see Table~\ref{tbl:Baseline models}). 
The second enhancement that is based on the inception layers is also beneficial compared with the basic UNET as it decreases the average MAE to $2.64$ with a kernel size set of $(3,5,7)$. 
Incorporating both strided convolutions and inception modules in the UNET leads to a lower average MAE of 2.20dB obtained with a kernel size set of $(1,5,7)$. 
This model provides the best performance for the power coverage prediction.

\setlength{\tabcolsep}{6pt}
\renewcommand{\arraystretch}{1.3}
\begin{table*}[!ht]
\caption{{Performance of the UNET model variants reported for 5 repeats over $32\times 32$ frames}}
\label{tbl:UNET Variations}
\begin{center}
\scalebox{0.93}{
\begin{tabular}{ccccccc}
\toprule
\multirow{2}{*}{Model} &  \multirow{2}{*}{Kernel size} & \multicolumn{3}{c}{MAE (dB)} & \multirow{2}{*}{Time (hr)} & \multirow{2}{*}{\# of parameters}\\
 \cline{3-5}
 &  &  min & max & average & & \\
\toprule
\multicolumn{1}{c}{UNET + Strided}             &    3         & 2.28       &    2.77   & 2.39        &  \multicolumn{1}{c}{2.25}            & \multicolumn{1}{c}{34.16M}                     \\ \midrule
\midrule
\multicolumn{1}{c}{UNET + Inception}           & (1,3,5)     & 2.25  & 3.98 & 3.40   & \multicolumn{1}{c}{4.27}        & \multicolumn{1}{c}{130.22M}              \\ \midrule
\multicolumn{1}{c}{UNET + Inception}           & (1,5,7)     & 2.27  & 3.96 & 3.39   & \multicolumn{1}{c}{2.73}        & \multicolumn{1}{c}{130.22M}              \\ \midrule
\multicolumn{1}{c}{UNET + Inception}           & (3,5,7)     & 2.27  & 3.39 & 2.64   & \multicolumn{1}{c}{3.95}        & \multicolumn{1}{c}{130.22M}              \\ \midrule
\midrule
\multicolumn{1}{c}{UNET + Strided + Inception} & (1,3,5)     & 2.24  & 3.84 & 2.78   & \multicolumn{1}{c}{2.69}        & \multicolumn{1}{c}{90.88M}               \\ \midrule
\multicolumn{1}{c}{\textbf{UNET + Strided + Inception}} & \textbf{(1,5,7)}     & \textbf{2.18}  & \textbf{2.22} & \textbf{2.20}   & \multicolumn{1}{c}{\textbf{5.08}}        & \multicolumn{1}{c}{\textbf{174.44M}}              \\ \midrule
\multicolumn{1}{c}{UNET + Strided + Inception} & (3,5,7)     & 2.22  & 2.31 & 2.26   & \multicolumn{1}{c}{5.77}        & \multicolumn{1}{c}{191.16M}              \\ \midrule
\end{tabular}
}
\end{center}
\end{table*}

\subsection{Performances of complex UNET models for large frame sizes}
We first compared the performance of different UNET architectures on datasets with different resolutions. The results of the experiments are reported in Table~\ref{tbl:summaryRess}. For the 32$\times$32 and 64$\times$64 frames, the best performing model is the UNET-SI-37. The average denormalized Mean Absolute Error (MAE) is 1.888dB and 2.410dB respectively. For both frame sizes, increasing the model complexity is not beneficial and leads to overfitting. We also visually examined the predictions and observed that there exists instances where one particular model performs better than others, while its overall performance (e.g., its average MAE) is worse.

\setlength{\tabcolsep}{6pt}
\renewcommand{\arraystretch}{1.15}
\begin{table*}[!ht]
\begin{center}
\caption{{Models performance comparison on different frame sizes and datasets over 5 repeats 
}}
\label{tbl:summaryRess}
\scalebox{0.81}{
\begin{tabular}{cccccccccc}
\toprule
\multirow{2}{*}{\textbf{Size}} &  \multirow{2}{*}{\textbf{Model}} &  \multirow{2}{*}{\textbf{Kernel Size}} & \multirow{2}{*}{\textbf{Areas}} &  \multirow{2}{*}{}&
\multicolumn{3}{c}{\textbf{denormalized MAE (dB)}}& \multirow{2}{*}{\textbf{Training time (hr)}} \\
 \cline{6-8}
& &  &  & & \textbf{Min} & \textbf{Max} & \textbf{Average} & \\
\toprule
\multirow{3}{*}{$32\times 32$} & \multicolumn{1}{c}{\textbf{UNET-SI-37}} & \textbf{(1,3,5)} & \textbf{5} &   & \textbf{0.002} & \textbf{15.373}   & \textbf{1.888} & \multicolumn{1}{c}{\textbf{0.2}} \\ 
& \multicolumn{1}{c}{UNET-SI-65} & (1,3,5)& 5 & &0.004 &28.150  &2.160& \multicolumn{1}{c}{2.6} \\ 
& \multicolumn{1}{c}{UNET-SI-91} & (1,3,5)& 5 & &0.000  &26.035&2.033& \multicolumn{1}{c}{1.3} \\ 
\midrule
\multirow{3}{*}{$64\times 64$} & \multicolumn{1}{c}{\textbf{UNET-SI-37}} & \textbf{(1,3,5)} & \textbf{5} & & \textbf{0.001} & \textbf{18.649} & \textbf{2.410} & \multicolumn{1}{c}{\textbf{4.2}}\\ 
& \multicolumn{1}{c}{UNET-SI-65} & (1,3,5)& 5 & &0.000   &17.756&2.490& \multicolumn{1}{c}{4.5}\\ 
& \multicolumn{1}{c}{UNET-SI-91} & (1,3,5)& 5 &  &0.000 &14.489&2.673   & \multicolumn{1}{c}{9.2}\\ 
\midrule
\multirow{3}{*}{$128\times 128$} & \multicolumn{1}{c}{UNET-SI-37} & (1,3,5)& 5 & & 0.000&14.073& 3.503& \multicolumn{1}{c}{3.8} \\ 
& \multicolumn{1}{c}{UNET-SI-65} & (1,3,5)& 5 & &0.001   &8.632&3.560& \multicolumn{1}{c}{11.4}\\ 
& \multicolumn{1}{c}{\textbf{UNET-SI-91}} & \textbf{(1,3,5)} & \textbf{5} & & \textbf{0.000} & \textbf{7.239} & \textbf{3.367} &\multicolumn{1}{c}{\textbf{13.1}} \\ 
\midrule
\multirow{5}{*}{$256\times 256$} & \multicolumn{1}{c}{UNET-SI-37} & (1,3,5)&5&&0.000&12.192&8.091& \multicolumn{1}{c}{10}\\
& \multicolumn{1}{c}{UNET-SI-65} & (1,3,5)& 5 & &0.000  &7.256 &3.749   & \multicolumn{1}{c}{25}\\ 
& \multicolumn{1}{c}{\textbf{UNET-SI 73}} & \textbf{(1,3,5)} & \textbf{5} &  & \textbf{0.000} & \textbf{6.020}  & \textbf{2.597} & \multicolumn{1}{c}{\textbf{13.3}}\\ 
 & \multicolumn{1}{c}{RadioUNETc Keras} & (3,4,5,6)&4&&0.012&15.599&8.488& \multicolumn{1}{c}{0.8} \\
& \multicolumn{1}{c}{RadioUNETc PyTorch} & (3,4,5,6)&4&&6.501&12.364&9.579& \multicolumn{1}{c}{4}\\
\bottomrule
\end{tabular}

}
\end{center}
\end{table*}

For higher frame sizes of 128$\times$128 and 256$\times$256, complex variants of the UNET performed better. Additionally, the difference in performance between the compared models increase moving from $128\times128$ to 256$\times$256 which suggests that the smaller models are underfitting for 256$\times$256 frames. The increased number of 2D convolutional layers and the triple-layer inception blocks in the UNET-SI-91 yielded the best denormalized MAE for 128$\times$128 which was around 3.367dB. However, the same architecture with 256$\times$256 frames required a memory size that was beyond the limits of the GPU we used in training, hence this frame size was tested on the UNET-SI 73 with fewer convolutional layers and two-layer inception modules. This structure decreased the MAE to 2.597dB and was the best performing model on 256$\times$256 in comparison with the basic and intermediate models UNET-SI-37 and UNET-SI-65.   

We trained the RadioUNETc architecture proposed by Levie et al.~\citep{levie2021radiounet} on 256$\times$256 frames. The models were trained in two different ways. In the first one, we converted the Pytorch implementation that the authors made publicly available to its Keras equivalent, then we trained this Keras implementation using the same hyperparameters (e.g., batch size and learning rate) we used for the UNET-SI models. The original RadioUNET PyTorch implementation contains small differences compared to our codebase written in Keras. The PyTorch implementation uses a different batch size and a learning rate scheduler. As we cannot discard the possibility of a subtle error during model conversion, we used a second method to train the RadioUNET that consists in implementing a dataloader in PyTorch to load 256$\times$256 frames and tested the model using the author's implementation with the same hyperparameters. The results in Table~\ref{tbl:summaryRess} show that both Keras and Pytorch implementations underperformed compared to the other UNET-SI architectures tested above. 

We also experimented with RadioMapSeer~\citep{levie2021radiounet} and compared different UNET variants. Our results show that UNET-SI-73 (i.e., the largest UNET model) provides better MAE values compared to others, which is consistent with our findings over the Wireless Insite dataset.
That is, large frame sizes benefit from deeper UNET architectures.
For this dataset, RadioUNET is able to overperform only the UNET-SI-37 (i.e., the smallest UNET model).
We observe that improved performance of large UNET models comes at the expense of significant increase in training times (e.g., 3 hours on average for RadioUNETc vs 52 hours on average for UNET-SI-73). 
While it is not reported in Table~\ref{tbl:summaryRess}, we also compared the performances of these four models using average RMSE values, which were 0.0469, 0.0176, 0.0158 and 0.0200 for UNET-SI-37, UNET-SI-65, UNET-SI-73, and RadioUNETc, respectively.
We note that these RMSE values are largely overlaps with the reported values in Levie et al.~\citep{levie2021radiounet}'s study.
In addition, we observe that overall MAE values are better for RadioMapSeer dataset when compared to $256\times 256$ frames from Wireless Insite dataset, which indicates those to be easier for coverage prediction.
That is, since the Wireless Insite dataset is more complex (generated using 3D simulations on detailed 3D maps, and the simulations contain higher number of reflections), the models trained on this dataset have slightly higher error compared to the RadioMapSeer dataset.

Overall, our analysis sheds light on how simpler models achieve best performance on lower resolution frames and have less computational complexity such as UNET-SI-37 on 32$\times$32 dataset with a runtime of 0.2 hours. Note that the performance of models across different dataset sizes are not comparable because of different test regions. 

\subsection{Visualizing the UNET predictions}
\begin{figure*}[!ht]
    \centering
    \subfloat[Building and transmitter location ]{\includegraphics[width=0.25\textwidth]{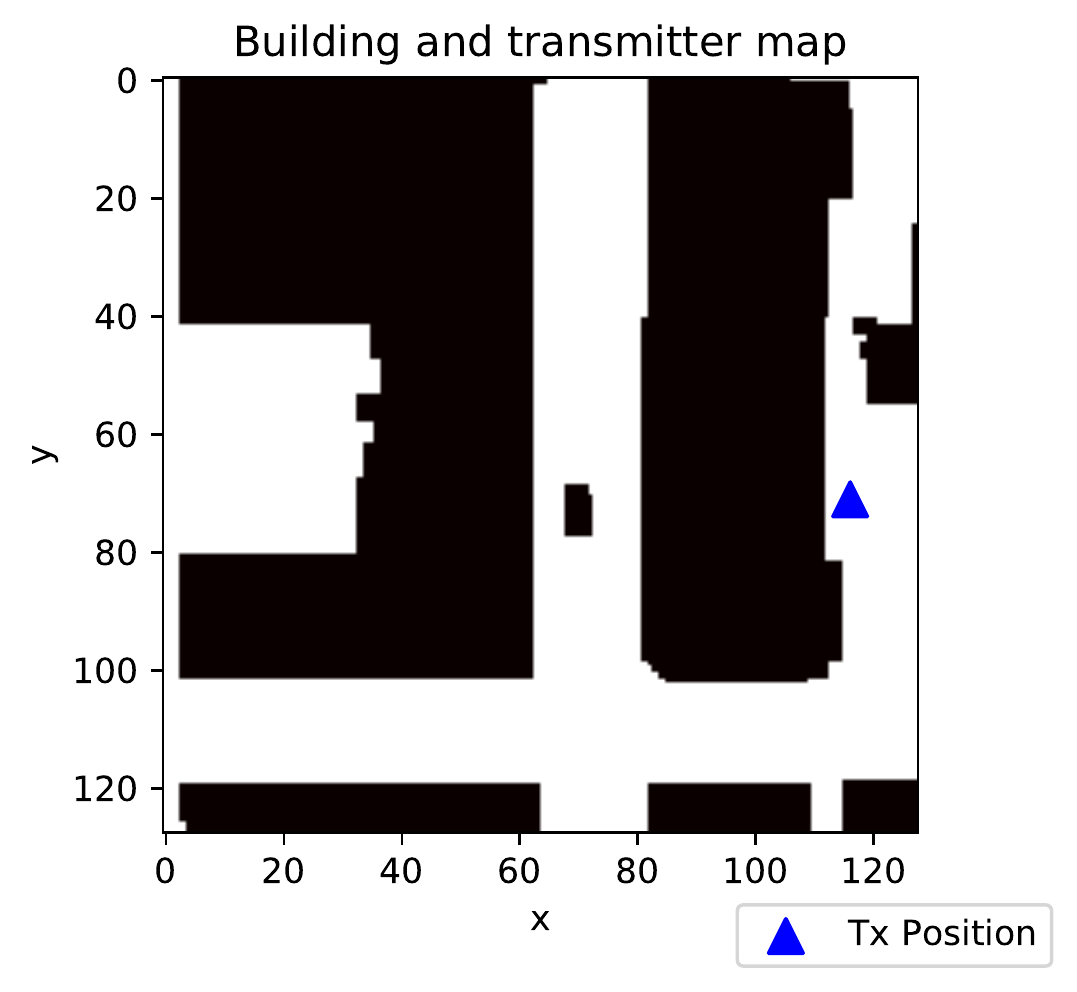}
    \label{fig:Building_128_200}
    }
    \qquad
    \subfloat[UNET-SI-37 prediction ]{\includegraphics[width=0.75\textwidth]{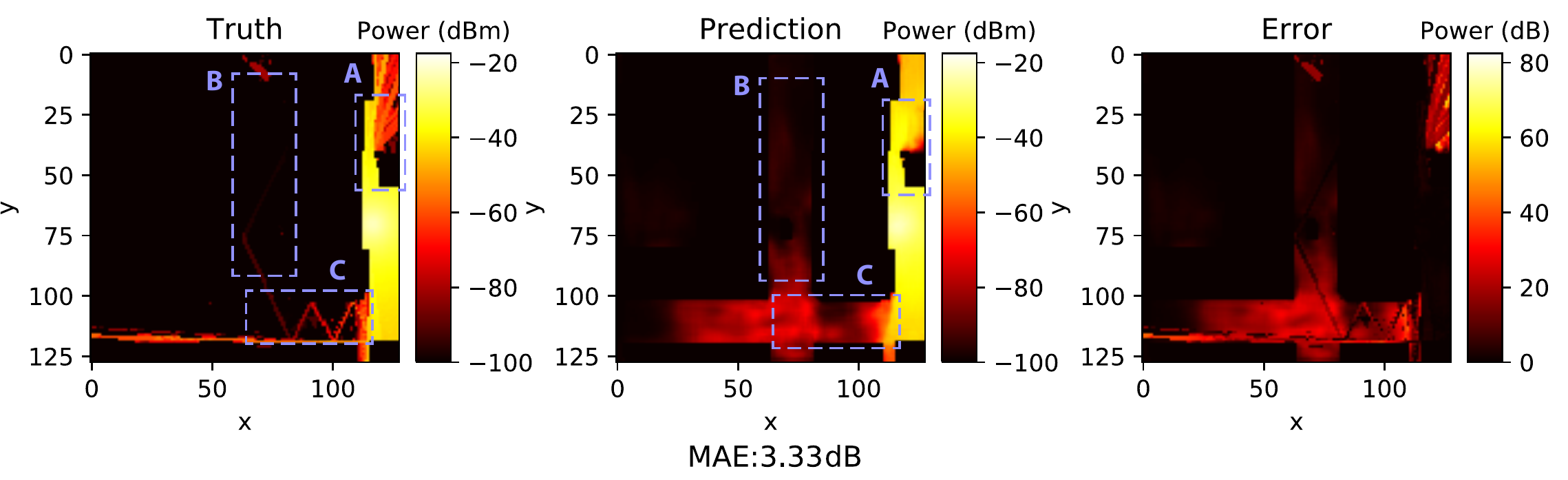}
    \label{fig:128_200_UNET prediction}
    }
    \qquad
     \subfloat[UNET-SI-65 prediction]{\includegraphics[width=0.75\textwidth]{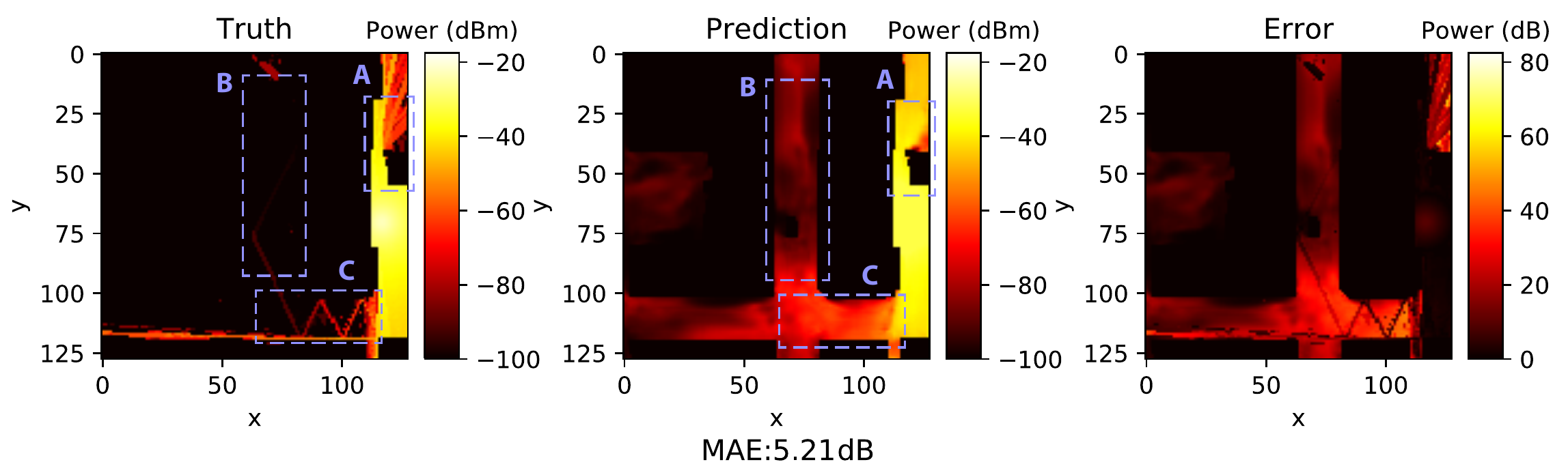}
    \label{fig:128_200_UNET 65 prediction}
    }
    \qquad
     \subfloat[UNET-SI-91 prediction]{\includegraphics[width=0.75\textwidth]{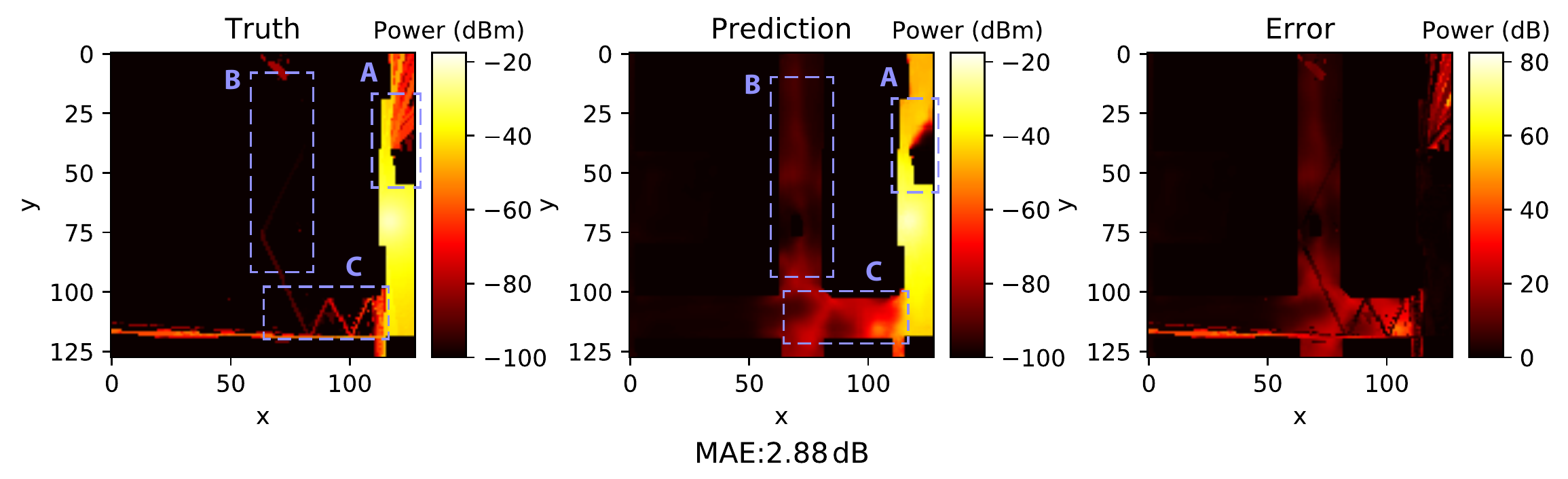}
    \label{fig:128_200_UNET 91 prediction}
    }
    \caption{Different UNET models predictions on 128$\times$128 frames}
    \label{fig:predictions_128_200}
\end{figure*}

\begin{figure*}[!ht]
    \centering
    \subfloat[Building and transmitter location ]{\includegraphics[width=0.25\textwidth]{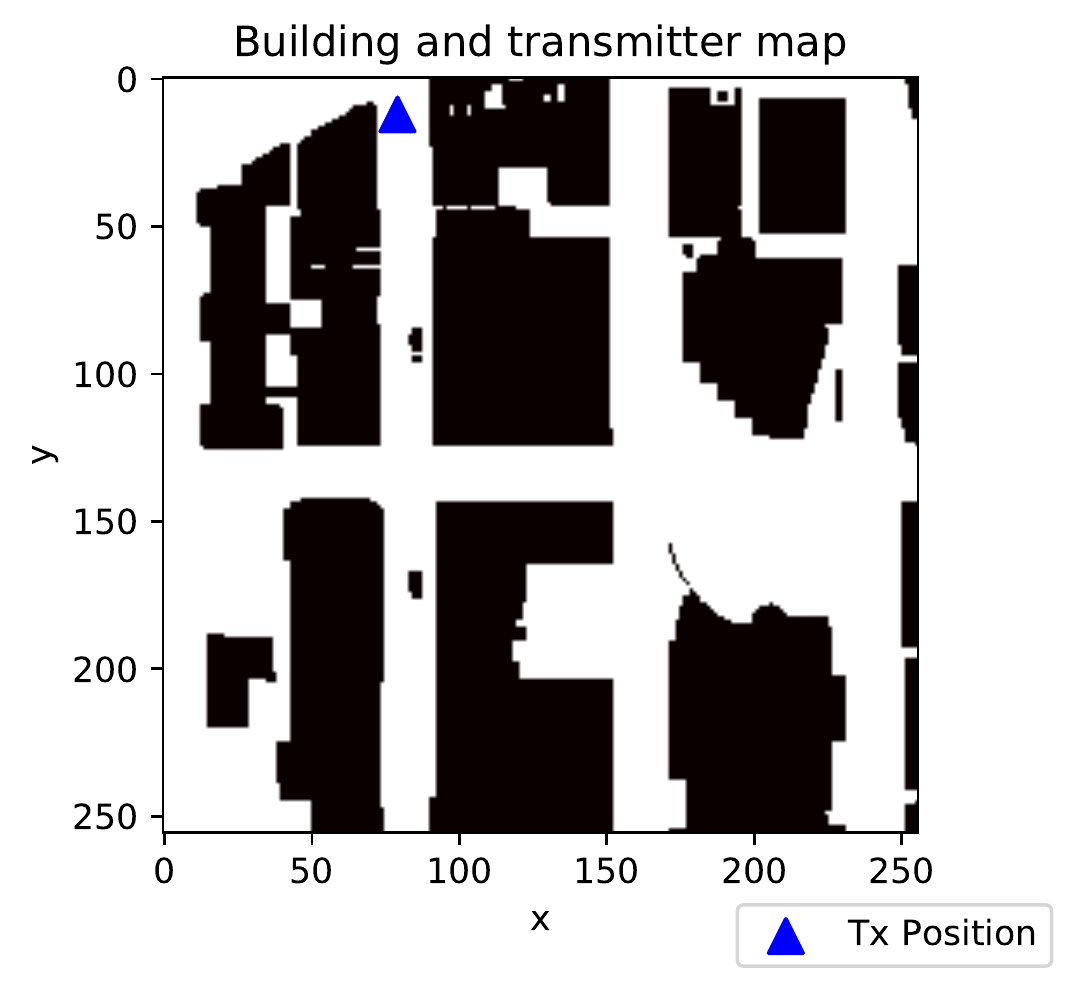}
    \label{fig:Building_256_879}
    }
    \qquad
    \subfloat[UNET-SI-37 prediction ]{\includegraphics[width=0.75\textwidth]{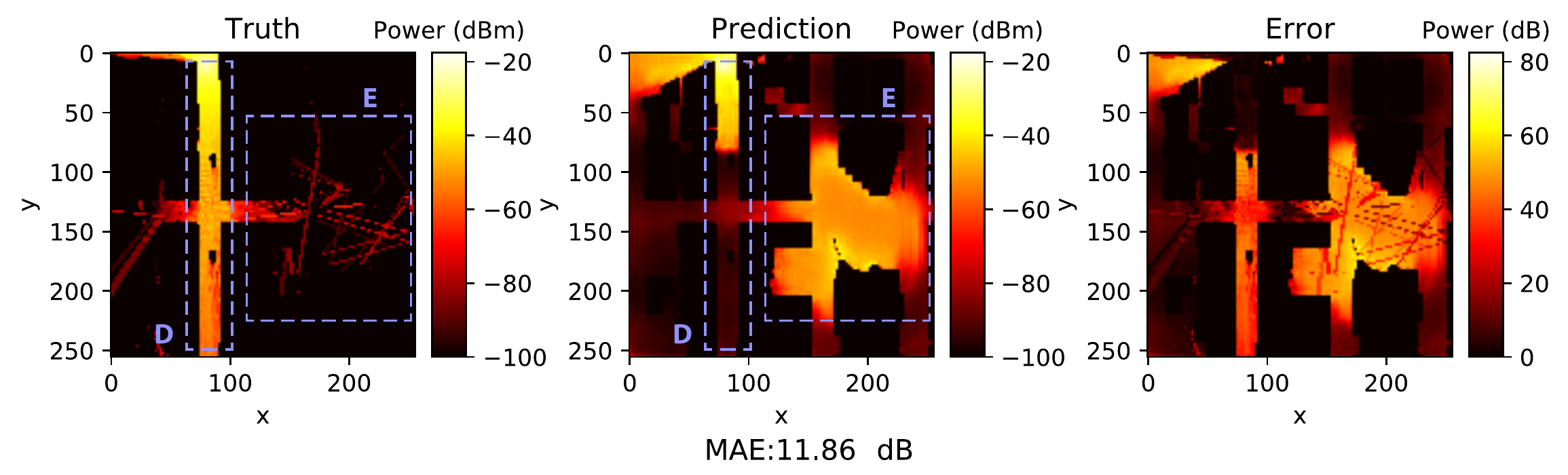}
    \label{fig:256_879_UNET prediction}
    }
    \qquad
     \subfloat[UNET-SI-65 prediction]{\includegraphics[width=0.75\textwidth]{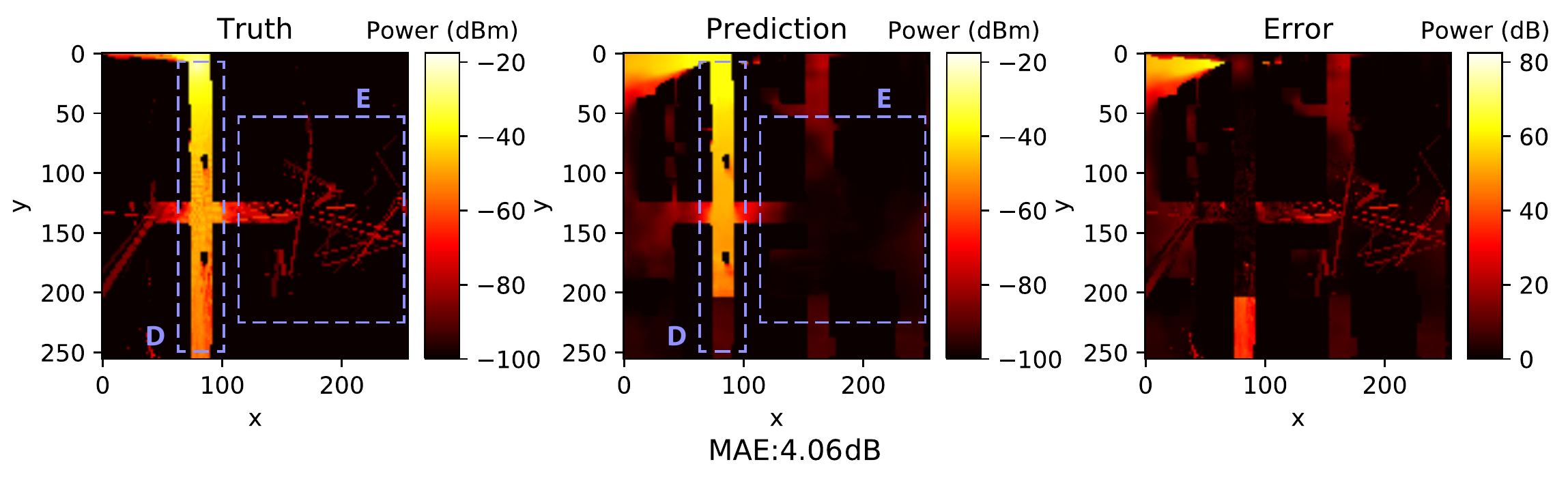}
    \label{fig:256_879_UNET 65 prediction}
    }
    \qquad
     \subfloat[UNET-SI 73 prediction]{\includegraphics[width=0.75\textwidth]{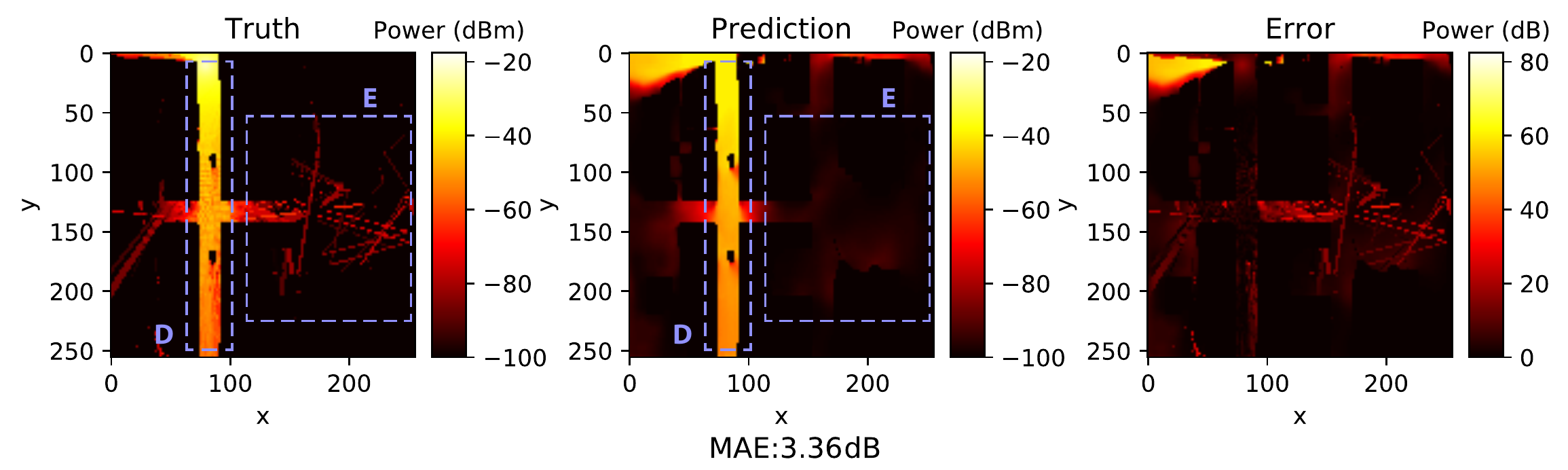}
    \label{fig:256_879_UNET 91 prediction}
    }
    \caption{Different UNET models predictions on 256$\times$256 frames }
    \label{fig:predictions_256_879}
\end{figure*}

A visual comparison between different UNET models on random samples is demonstrated in Fig.~\ref{fig:predictions_128_200} and Fig.~\ref{fig:predictions_256_879}.
Specifically, Fig.~\ref{fig:predictions_128_200} shows the prediction of the UNET-SI-37, 65 and 91 on a $128\times128$ frame. The smallest error is obtained with the complex model UNET-SI-91. The differences between models can be seen on the three areas annotated A, B and C. The power coverage predicted by the UNET-SI-91 in area A of Fig.~\ref{fig:128_200_UNET 91 prediction} was the most similar to the ground truth. In the central region B, this model predicted the lowest energy distribution compared to the two other models. Area C is interesting as it shows that the complex model UNET-SI-91 predicted an energy pattern with alternate highs and lows. These patterns resemble their equivalent in the ground truth, though not at the same level of detail, but at least far better than the two other models UNET-SI-37 and 65 in Fig.~\ref{fig:128_200_UNET prediction} and Fig.~\ref{fig:128_200_UNET 65 prediction}, respectively. This particular example shows the benefit of a more complex inception structure in capturing propagation patterns such as the ones in area C.

Fig.~\ref{fig:predictions_256_879} shows the predictions on a random frame of size $256\times256$. The prediction by the simplest model UNET-SI-37 depicted in Fig.~\ref{fig:predictions_256_879}b was less consistent with the ground truth and produced a high energy region in area E that is nonexistent in the truth. Furthermore, the power prediction seems to be cut in area D because of an insufficient number of convolutional layers. This abrupt cut in area D is less noticeable with the UNET-SI-65 in Fig.~\ref{fig:256_879_UNET 65 prediction} and completely absent with the UNET-SI 73 prediction in Fig.~\ref{fig:256_879_UNET 91 prediction}. The prediction with this model is very similar to the truth. This comparison illustrates the benefit of an increased number of convolutional layers and inception modules to the learning process.

Though the UNET-SI 73 performed relatively well, we still note that it was far from providing the fine-grained reflections observed in the ground truth image. The absence of these sharp propagation patterns is possibly due to the loss of details with strided convolutions as well as the few long skip connections between the encoder and decoder in the UNET structure. However, such predictions could be very useful to get a real-time estimate of the overall distribution of power while looking for the best transmitter locations in a complex urban setting.

\subsection{UNET models in practice}
We used the UNET-SI-37, the best performing model on 32$\times$32 frames, to create an interactive interface in which the designer can add/remove building pixels and transmitters by left/right clicking the image respectively. After each interaction, the model predicts the power coverage in a few tens of milliseconds.

\begin{figure*}[!ht]
    \centering
        \subfloat[One transmitter ]{\includegraphics[width=0.28\textwidth]{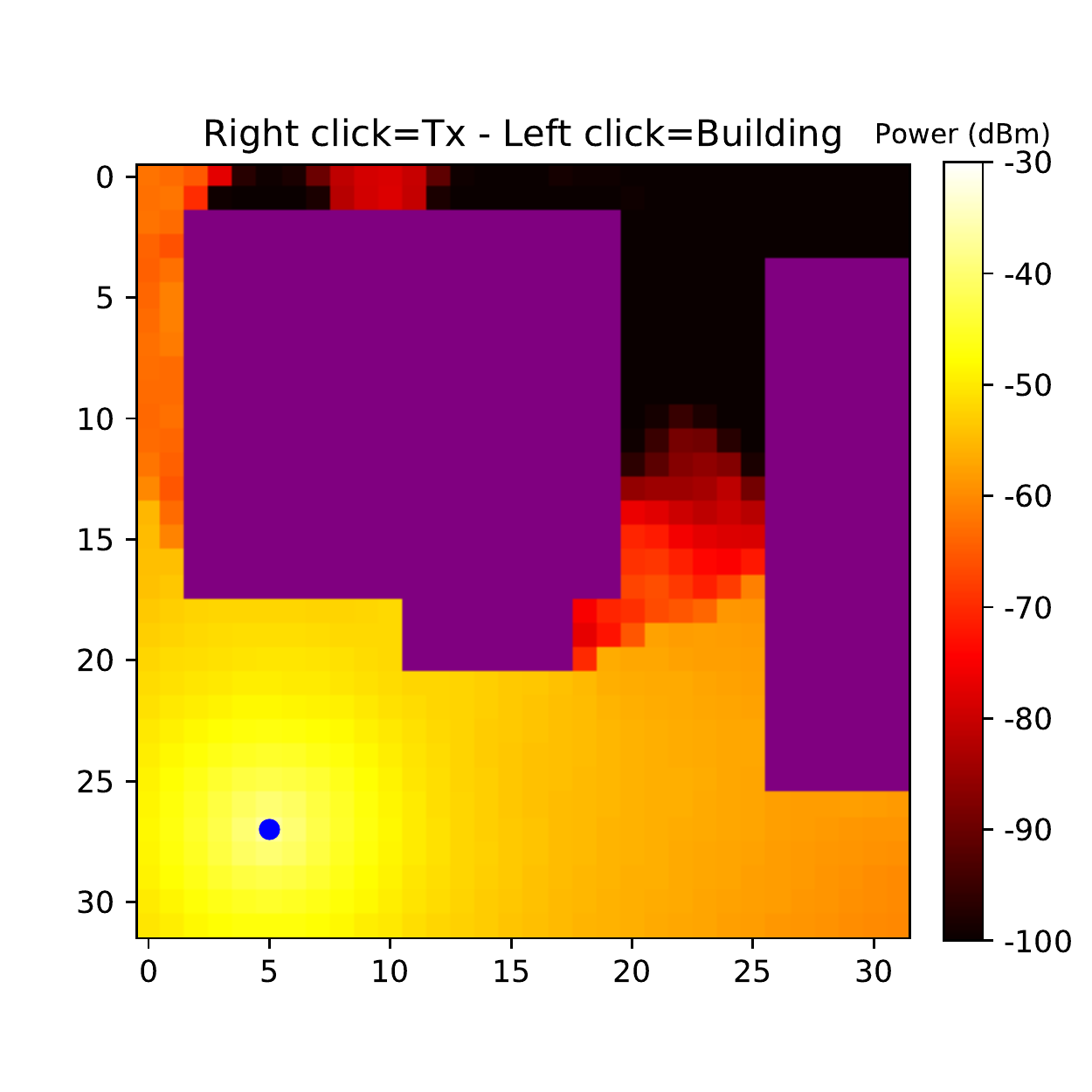}
        \label{fig:frame_1Tx}
        }
        \qquad
        \subfloat[Two transmitters ]{\includegraphics[width=0.28\textwidth]{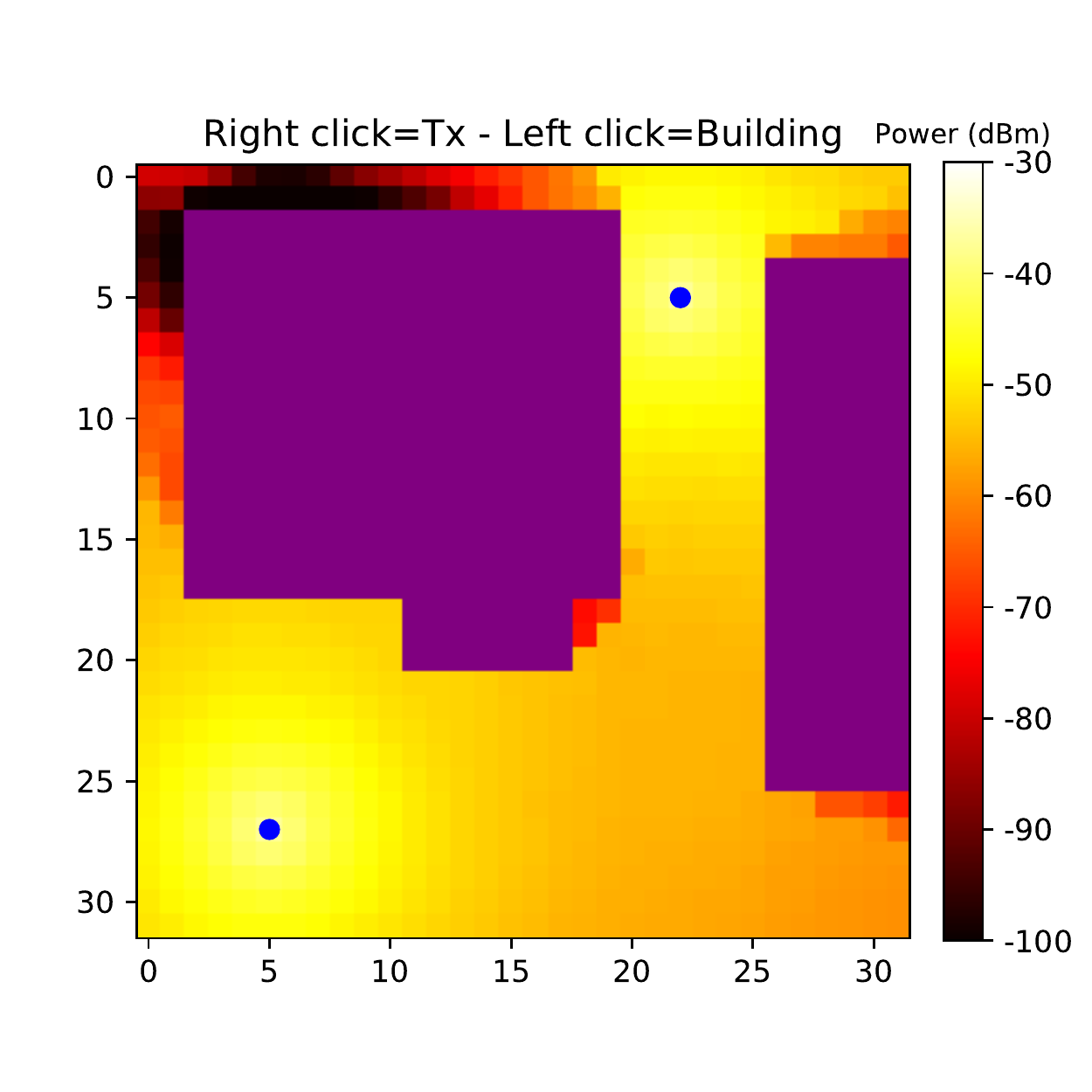}
        \label{fig:frame_2Tx}
        }
    \caption{Experiment with multiple transmitters using the interactive interface. Buildings are added manually and shown in purple. The blue dots indicate arbitrary transmitter locations. The model was only trained with one transmitter as in (a). When adding a second transmitter with the interactive interface (b), the power coverage provided by the model seems to be coherent.}
    \label{fig:multi_Tx}
\end{figure*}

\begin{figure*}[!ht]
    \centering
        \subfloat[Moving object Pos.1 ]{\includegraphics[width=0.21\textwidth]{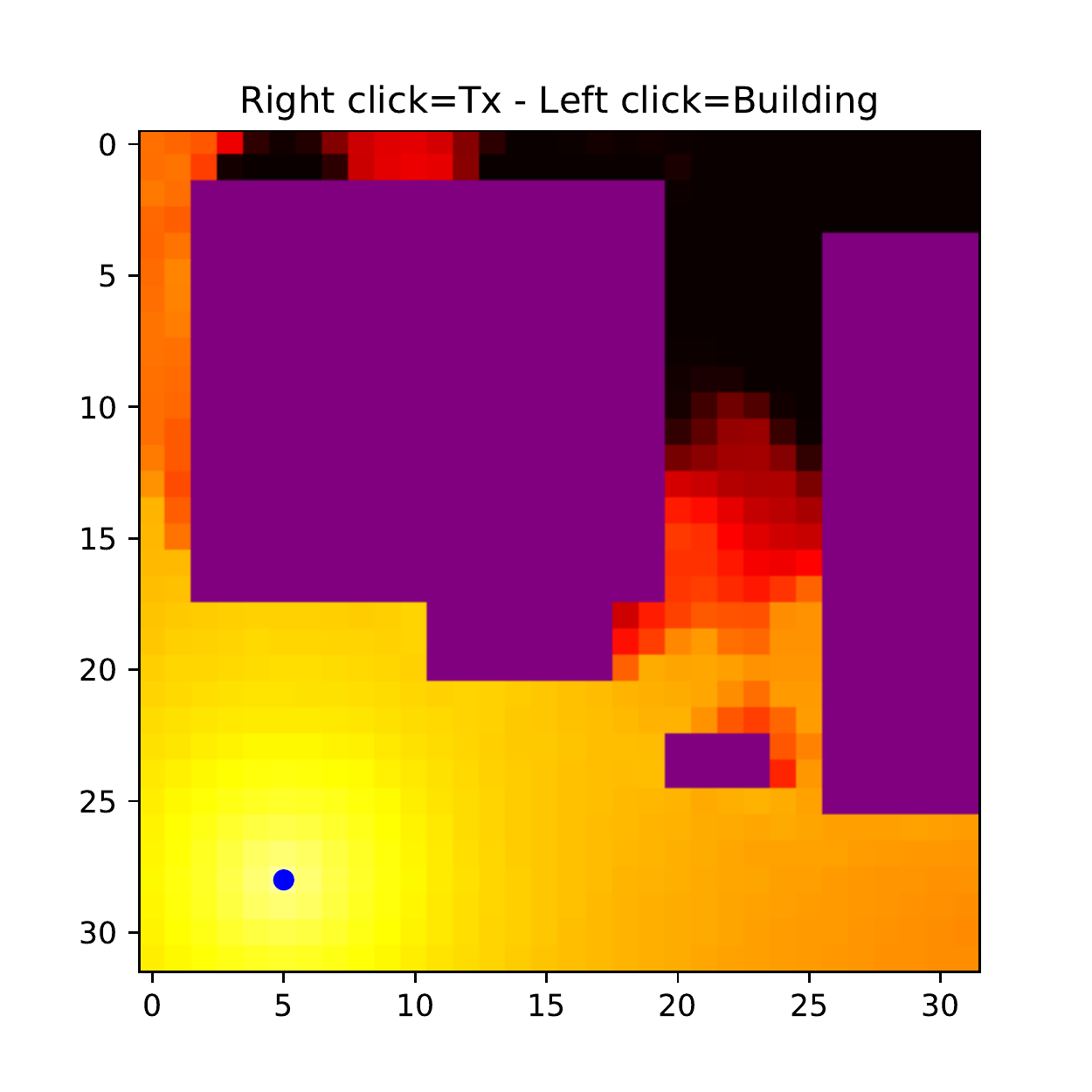}
        \label{fig:frame_Mo_7}
        }
        \qquad
        \subfloat[Moving object Pos.2 ]{\includegraphics[width=0.21\textwidth]{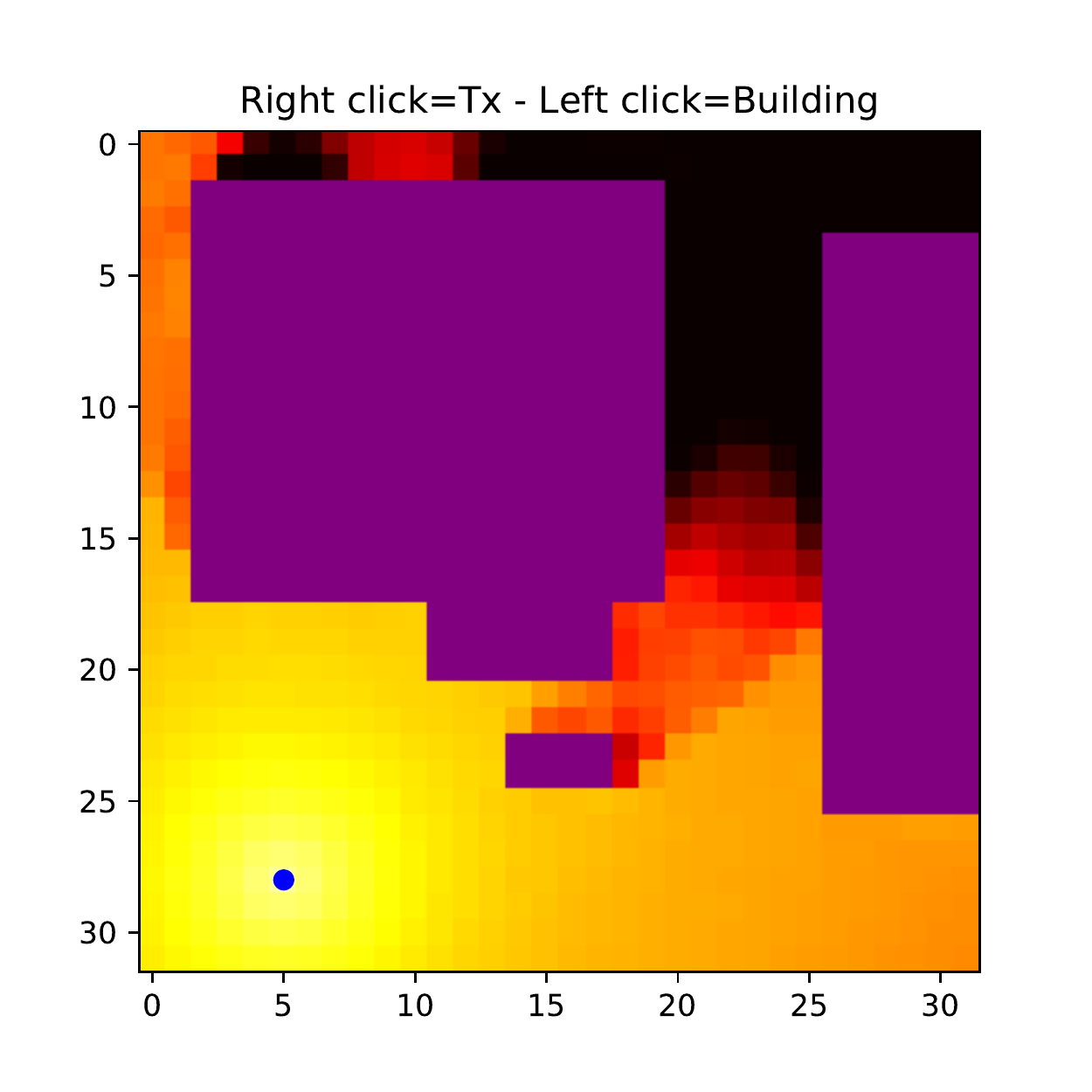}
        \label{fig:frame_Mo_5}
        }
        \subfloat[Moving object Pos.3 ]{\includegraphics[width=0.21\textwidth]{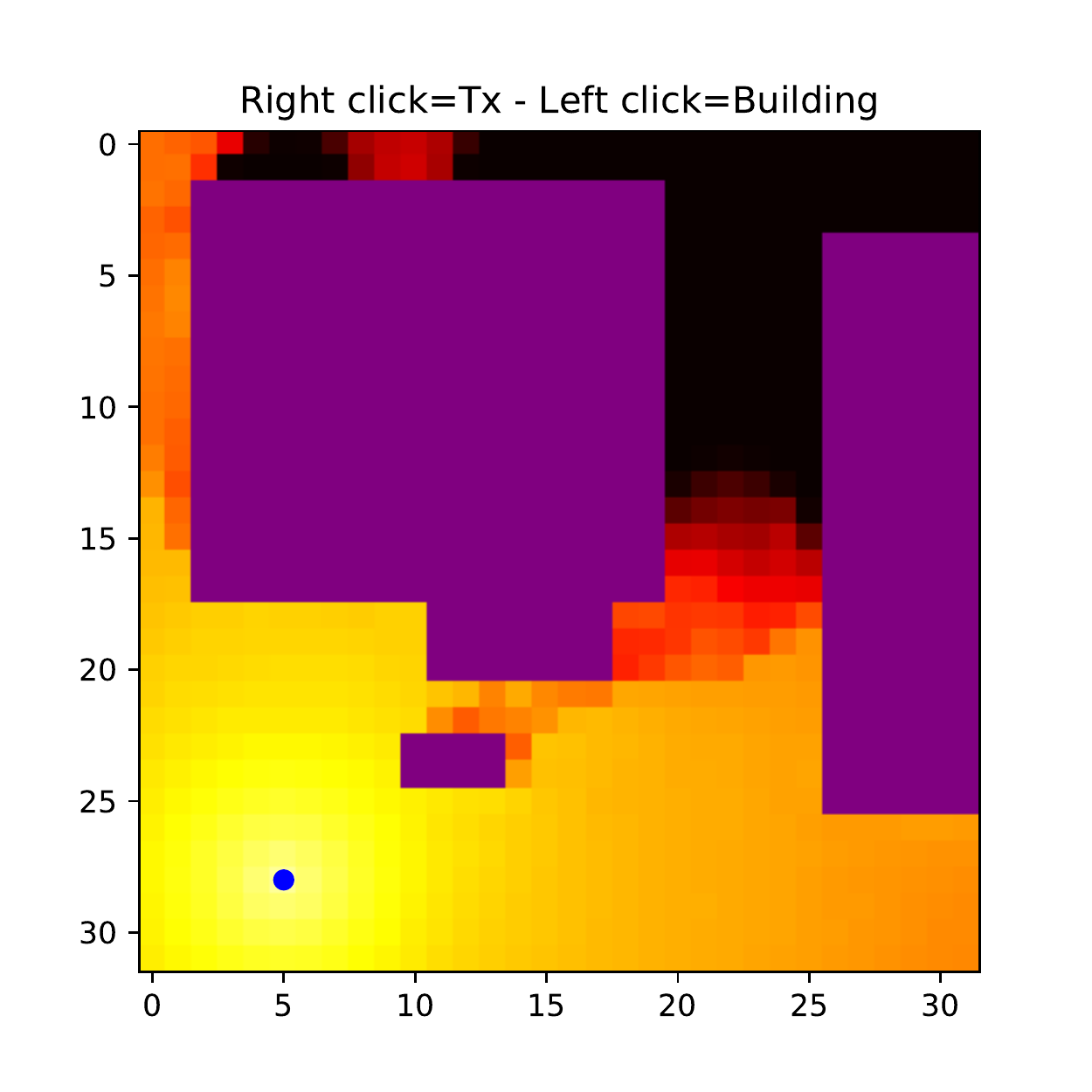}
        \label{fig:frame_Mo_3}
        }
        \subfloat[Moving object Pos.4 ]{\includegraphics[width=0.21\textwidth]{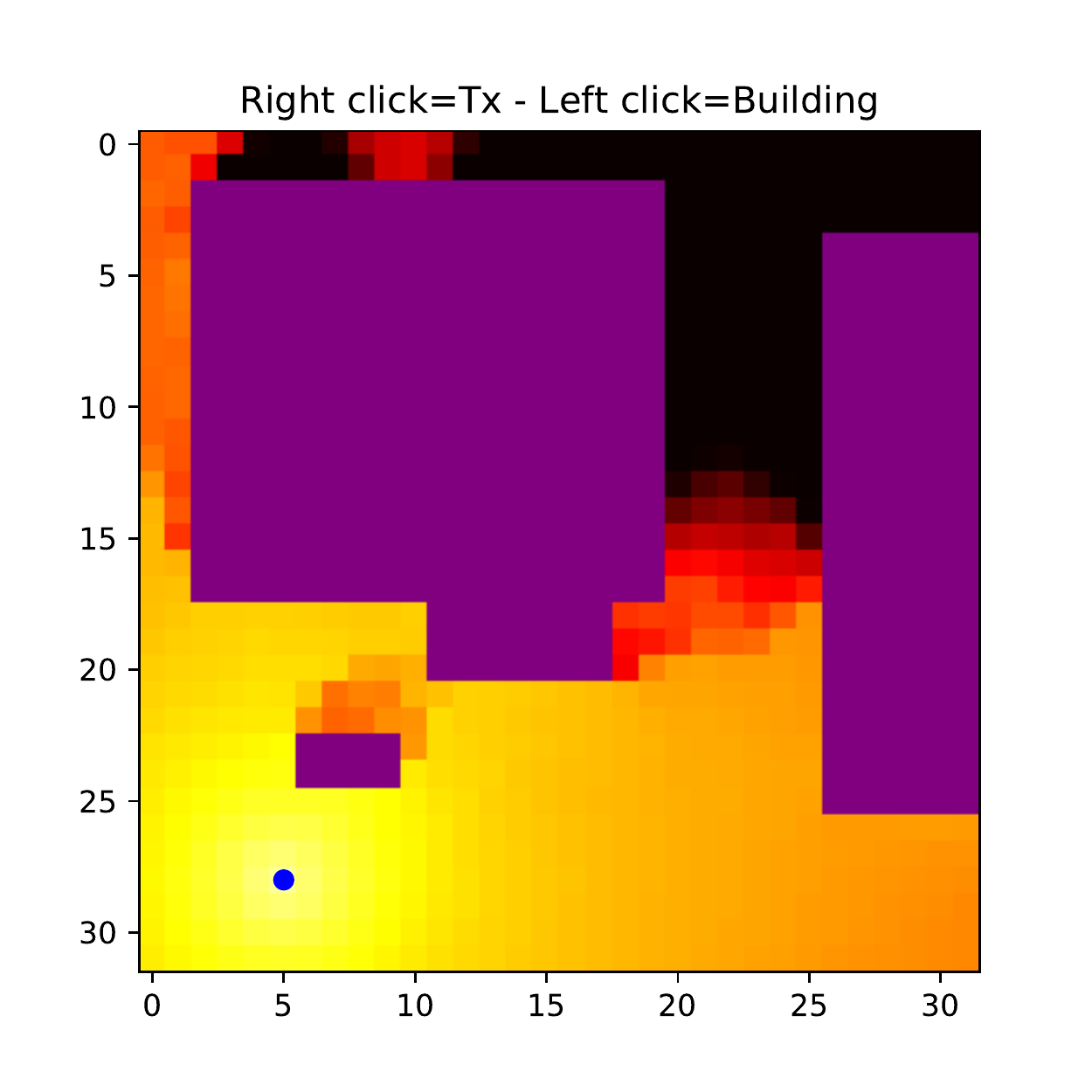}
        \label{fig:frame_Mo_1}
        }
    \caption{Experiment with a moving object using the interactive interface. Purple blocks are buildings drawn manually. The transmitter is shown as a blue dot. The four frames show the movement sequence of an object from right to left. The model predicts instantaneously the change in power coverage in each frame. The coverage seems to be consistent with radio propagation rules.}
    \label{fig:moving_object}
\end{figure*}

Fig.~\ref{fig:multi_Tx} and Fig.~\ref{fig:moving_object} show two applications of this interactive interface. Buildings, shown in purple, are created manually or imported from the dataset. The transmitter is depicted as a blue dot. Fig.~\ref{fig:multi_Tx} shows an example with one and two transmitters. It is important to note that even though the training consisted in frames with one transmitter only, we see that predictions appear reasonable when adding a second transmitter. In Fig.~\ref{fig:frame_1Tx}, the dark colors between the two buildings shows that very little radio frequency power reaches this space from the transmitter. In Fig.~\ref{fig:frame_2Tx}, we note that a second transmitter placed between the two buildings significantly increases the power coverage in this region. The overall power prediction seems to consider both transmitters even if this multi-transmitter setting is entirely new to the model.

Fig.~\ref{fig:moving_object} illustrates a second application of real-time predictions where we study the impact of a moving object on power distribution. The prediction can be obtained in two ways, either by drawing the object manually in every position or by generating a sequence of images with different locations of the object, then feeding the image sequence to the model. Fig.~\ref{fig:moving_object} shows the model predictions from this sequence. As the object moves from right to left, the change in power values can be observed by looking at the shape of the shadow cast by the moving object that seems congruent with the propagation principles.


\section{Conclusions}\label{sec:conclusions}
In this paper, we investigated the effectiveness of CNN and UNET models for radio signal prediction in large urban environments. 
We considered various modifications such as increasing the complexity by adding more convolutional layers and deeper inception layers. These proposed models have reasonable error rates with high resolution frames. They also performed better compared to the RadioUNET model proposed by Levie et al.~\citep{levie2021radiounet}. In particular, our empirical work showed how fine-tuned deep learning models can be used for more complex datasets. These models can predict the power coverage instantly after training and because of their generalizability, they can be tested on different urban regions.
 
There are certain limitations of our work. 
First, 
while we conduct a detailed empirical study with distinct datasets,
it would be important to test out the further generalizability of our model using datasets obtained from different sources. 
In addition, the performances of our UNET models can be further improved by using more complex inception blocks and higher number of convolution layers. 
However, our computing environment does not allow such UNET architectures, which necessitates investigation of more memory efficient deep learning models for the power prediction task. 
In addition to exploring ways to address these issues, we also plan to mitigate the issue of reflection prediction in images as a future research direction. 
A possible solution may be the introduction of short skip connections with residual blocks as described by Drozdzal et al.~\citep{drozdzal2016importance} or dense blocks as proposed by Jegou et al.~\citep{jegou2017one}.

\section*{Acknowledgments}
This research is in part supported by Communications Research Centre (CRC) Canada. 


\bibliographystyle{plainnat}
\bibliography{refs}

\begin{thebibliography}{30}
\providecommand{\natexlab}[1]{#1}
\providecommand{\url}[1]{\texttt{#1}}
\expandafter\ifx\csname urlstyle\endcsname\relax
  \providecommand{\doi}[1]{doi: #1}\else
  \providecommand{\doi}{doi: \begingroup \urlstyle{rm}\Url}\fi

\bibitem[Angeles and Dadios(2015)]{angeles2015neural}
Joel C~Delos Angeles and Elmer~P Dadios.
\newblock Neural network-based path loss prediction for digital tv macrocells.
\newblock In \emph{2015 International Conference on Humanoid, Nanotechnology,
  Information Technology, Communication and Control, Environment and Management
  (HNICEM)}, pages 1--9. IEEE, 2015.

\bibitem[{\c{C}}i{\c{c}}ek et~al.(2016){\c{C}}i{\c{c}}ek, Abdulkadir, Lienkamp,
  Brox, and Ronneberger]{cciccek20163d}
{\"O}zg{\"u}n {\c{C}}i{\c{c}}ek, Ahmed Abdulkadir, Soeren~S Lienkamp, Thomas
  Brox, and Olaf Ronneberger.
\newblock 3d u-net: learning dense volumetric segmentation from sparse
  annotation.
\newblock In \emph{International conference on medical image computing and
  computer-assisted intervention}, pages 424--432. Springer, 2016.

\bibitem[Ciresan et~al.(2012)Ciresan, Giusti, Gambardella, and
  Schmidhuber]{ciresan2012deep}
Dan Ciresan, Alessandro Giusti, Luca~M Gambardella, and J{\"u}rgen Schmidhuber.
\newblock Deep neural networks segment neuronal membranes in electron
  microscopy images.
\newblock In \emph{Advances in neural information processing systems}, pages
  2843--2851, 2012.

\bibitem[Drozdzal et~al.(2016)Drozdzal, Vorontsov, Chartrand, Kadoury, and
  Pal]{drozdzal2016importance}
Michal Drozdzal, Eugene Vorontsov, Gabriel Chartrand, Samuel Kadoury, and Chris
  Pal.
\newblock The importance of skip connections in biomedical image segmentation.
\newblock In \emph{Deep Learning and Data Labeling for Medical Applications},
  pages 179--187. Springer, 2016.

\bibitem[Erceg et~al.(1999)Erceg, Greenstein, Tjandra, Parkoff, Gupta, Kulic,
  Julius, and Bianchi]{erceg1999empirically}
Vinko Erceg, Larry~J Greenstein, Sony~Y Tjandra, Seth~R Parkoff, Ajay Gupta,
  Boris Kulic, Arthur~A Julius, and Renee Bianchi.
\newblock An empirically based path loss model for wireless channels in
  suburban environments.
\newblock \emph{IEEE Journal on selected areas in communications}, 17\penalty0
  (7):\penalty0 1205--1211, 1999.

\bibitem[Girshick et~al.(2015)Girshick, Donahue, Darrell, and
  Malik]{girshick2015region}
Ross Girshick, Jeff Donahue, Trevor Darrell, and Jitendra Malik.
\newblock Region-based convolutional networks for accurate object detection and
  segmentation.
\newblock \emph{IEEE transactions on pattern analysis and machine
  intelligence}, 38\penalty0 (1):\penalty0 142--158, 2015.

\bibitem[Gupta et~al.(2014)Gupta, Girshick, Arbel{\'a}ez, and
  Malik]{gupta2014learning}
Saurabh Gupta, Ross Girshick, Pablo Arbel{\'a}ez, and Jitendra Malik.
\newblock Learning rich features from rgb-d images for object detection and
  segmentation.
\newblock In \emph{European conference on computer vision}, pages 345--360.
  Springer, 2014.

\bibitem[Hata(1980)]{hata1980empirical}
Masaharu Hata.
\newblock Empirical formula for propagation loss in land mobile radio services.
\newblock \emph{IEEE transactions on Vehicular Technology}, 29\penalty0
  (3):\penalty0 317--325, 1980.

\bibitem[He et~al.(2015)He, Zhang, Ren, and Sun]{he2015spatial}
Kaiming He, Xiangyu Zhang, Shaoqing Ren, and Jian Sun.
\newblock Spatial pyramid pooling in deep convolutional networks for visual
  recognition.
\newblock \emph{IEEE transactions on pattern analysis and machine
  intelligence}, 37\penalty0 (9):\penalty0 1904--1916, 2015.

\bibitem[He et~al.(2016)He, Zhang, Ren, and Sun]{he2016deep}
Kaiming He, Xiangyu Zhang, Shaoqing Ren, and Jian Sun.
\newblock Deep residual learning for image recognition.
\newblock In \emph{Proceedings of the IEEE conference on computer vision and
  pattern recognition}, pages 770--778, 2016.

\bibitem[J{\'e}gou et~al.(2017)J{\'e}gou, Drozdzal, Vazquez, Romero, and
  Bengio]{jegou2017one}
Simon J{\'e}gou, Michal Drozdzal, David Vazquez, Adriana Romero, and Yoshua
  Bengio.
\newblock The one hundred layers tiramisu: Fully convolutional densenets for
  semantic segmentation.
\newblock In \emph{Proceedings of the IEEE conference on computer vision and
  pattern recognition workshops}, pages 11--19, 2017.

\bibitem[Kingma and Ba(2014)]{adam}
Diederik~P Kingma and Jimmy Ba.
\newblock Adam: A method for stochastic optimization.
\newblock \emph{arXiv preprint arXiv:1412.6980}, 2014.

\bibitem[Levie et~al.(2021)Levie, Yapar, Kutyniok, and
  Caire]{levie2021radiounet}
Ron Levie, {\c{C}}a{\u{g}}kan Yapar, Gitta Kutyniok, and Giuseppe Caire.
\newblock Radiounet: Fast radio map estimation with convolutional neural
  networks.
\newblock \emph{IEEE Transactions on Wireless Communications}, 2021.

\bibitem[Lim et~al.(2017)Lim, Son, Kim, Nah, and Mu~Lee]{lim2017enhanced}
Bee Lim, Sanghyun Son, Heewon Kim, Seungjun Nah, and Kyoung Mu~Lee.
\newblock Enhanced deep residual networks for single image super-resolution.
\newblock In \emph{Proceedings of the IEEE conference on computer vision and
  pattern recognition workshops}, pages 136--144, 2017.

\bibitem[Litjens et~al.(2017)Litjens, Kooi, Bejnordi, Setio, Ciompi,
  Ghafoorian, Van Der~Laak, Van~Ginneken, and S{\'a}nchez]{litjens2017survey}
Geert Litjens, Thijs Kooi, Babak~Ehteshami Bejnordi, Arnaud Arindra~Adiyoso
  Setio, Francesco Ciompi, Mohsen Ghafoorian, Jeroen~Awm Van Der~Laak, Bram
  Van~Ginneken, and Clara~I S{\'a}nchez.
\newblock A survey on deep learning in medical image analysis.
\newblock \emph{Medical image analysis}, 42:\penalty0 60--88, 2017.

\bibitem[Long et~al.(2015)Long, Shelhamer, and Darrell]{long2015fully}
Jonathan Long, Evan Shelhamer, and Trevor Darrell.
\newblock Fully convolutional networks for semantic segmentation.
\newblock In \emph{Proceedings of the IEEE conference on computer vision and
  pattern recognition}, pages 3431--3440, 2015.

\bibitem[Milletari et~al.(2016)Milletari, Navab, and Ahmadi]{milletari2016v}
Fausto Milletari, Nassir Navab, and Seyed-Ahmad Ahmadi.
\newblock V-net: Fully convolutional neural networks for volumetric medical
  image segmentation.
\newblock In \emph{2016 Fourth International Conference on 3D Vision (3DV)},
  pages 565--571. IEEE, 2016.

\bibitem[Mohammadjafari et~al.(2020)Mohammadjafari, Roginsky, Kavurmacioglu,
  Cevik, Ethier, and Bener]{mohammadjafari2020machine}
Sanaz Mohammadjafari, Sophie Roginsky, Emir Kavurmacioglu, Mucahit Cevik,
  Jonathan Ethier, and Ayse~Basar Bener.
\newblock Machine learning-based radio coverage prediction in urban
  environments.
\newblock \emph{IEEE Transactions on Network and Service Management},
  17\penalty0 (4):\penalty0 2117--2130, 2020.

\bibitem[Pinheiro and Collobert(2014)]{pinheiro2014recurrent}
Pedro~HO Pinheiro and Ronan Collobert.
\newblock Recurrent convolutional neural networks for scene labeling.
\newblock In \emph{31st International Conference on Machine Learning (ICML)},
  number CONF, 2014.

\bibitem[Radenovi{\'c} et~al.(2016)Radenovi{\'c}, Tolias, and
  Chum]{radenovic2016cnn}
Filip Radenovi{\'c}, Giorgos Tolias, and Ond{\v{r}}ej Chum.
\newblock Cnn image retrieval learns from bow: Unsupervised fine-tuning with
  hard examples.
\newblock In \emph{European conference on computer vision}, pages 3--20.
  Springer, 2016.

\bibitem[Radenovi{\'c} et~al.(2018)Radenovi{\'c}, Tolias, and
  Chum]{radenovic2018fine}
Filip Radenovi{\'c}, Giorgos Tolias, and Ond{\v{r}}ej Chum.
\newblock Fine-tuning cnn image retrieval with no human annotation.
\newblock \emph{IEEE transactions on pattern analysis and machine
  intelligence}, 41\penalty0 (7):\penalty0 1655--1668, 2018.

\bibitem[Rautiainen et~al.(2002)Rautiainen, Wolfle, and
  Hoppe]{rautiainen2002verifying}
Terhi Rautiainen, G~Wolfle, and Reiner Hoppe.
\newblock Verifying path loss and delay spread predictions of a 3d ray tracing
  propagation model in urban environment.
\newblock In \emph{Proceedings IEEE 56th Vehicular Technology Conference},
  pages 2470--2474. IEEE, 2002.

\bibitem[REMCOM(2020)]{insite}
REMCOM.
\newblock Wireless insite, v3.3.
\newblock \url{https://www.remcom.com/wireless-insite-em-propagation-software},
  2020.

\bibitem[Ronneberger et~al.(2015)Ronneberger, Fischer, and Brox]{ronn2015}
Olaf Ronneberger, Philipp Fischer, and Thomas Brox.
\newblock U-net: Convolutional networks for biomedical image segmentation.
\newblock In Nassir Navab, Joachim Hornegger, William~M. Wells, and
  Alejandro~F. Frangi, editors, \emph{Medical Image Computing and
  Computer-Assisted Intervention -- MICCAI 2015}, pages 234--241, Cham, 2015.
  Springer International Publishing.
\newblock ISBN 978-3-319-24574-4.

\bibitem[Simonyan and Zisserman(2014)]{simonyan2014very}
Karen Simonyan and Andrew Zisserman.
\newblock Very deep convolutional networks for large-scale image recognition.
\newblock \emph{arXiv preprint arXiv:1409.1556}, 2014.

\bibitem[Springenberg et~al.(2014)Springenberg, Dosovitskiy, Brox, and
  Riedmiller]{springenberg2014striving}
Jost~Tobias Springenberg, Alexey Dosovitskiy, Thomas Brox, and Martin
  Riedmiller.
\newblock Striving for simplicity: The all convolutional net.
\newblock \emph{arXiv preprint arXiv:1412.6806}, 2014.

\bibitem[Szegedy et~al.(2015)Szegedy, Liu, Jia, Sermanet, Reed, Anguelov,
  Erhan, Vanhoucke, and Rabinovich]{szegedy2015going}
Christian Szegedy, Wei Liu, Yangqing Jia, Pierre Sermanet, Scott Reed, Dragomir
  Anguelov, Dumitru Erhan, Vincent Vanhoucke, and Andrew Rabinovich.
\newblock Going deeper with convolutions.
\newblock In \emph{Proceedings of the IEEE conference on computer vision and
  pattern recognition}, pages 1--9, 2015.

\bibitem[Wahl et~al.(2005)Wahl, W{\"o}lfle, Wertz, Wildbolz, and
  Landstorfer]{wahl2005dominant}
Ren{\'e} Wahl, Gerd W{\"o}lfle, Philipp Wertz, Pascal Wildbolz, and Friedrich
  Landstorfer.
\newblock Dominant path prediction model for urban scenarios.
\newblock \emph{14th IST Mobile and Wireless Communications Summit, Dresden
  (Germany)}, 2005.

\bibitem[Yi et~al.(2017)Yi, Zhang, Tan, and Gong]{yi2017dualgan}
Zili Yi, Hao Zhang, Ping Tan, and Minglun Gong.
\newblock Dualgan: Unsupervised dual learning for image-to-image translation.
\newblock In \emph{Proceedings of the IEEE international conference on computer
  vision}, pages 2849--2857, 2017.

\bibitem[Zhao et~al.(2017)Zhao, Feng, Wu, and Yan]{zhao2017survey}
Bo~Zhao, Jiashi Feng, Xiao Wu, and Shuicheng Yan.
\newblock A survey on deep learning-based fine-grained object classification
  and semantic segmentation.
\newblock \emph{International Journal of Automation and Computing}, 14\penalty0
  (2):\penalty0 119--135, 2017.

\end{thebibliography}

\end{document}